\documentclass[pmlr,twocolumn,10pt]{jmlr} %

\usepackage{booktabs}
\usepackage{siunitx}

\usepackage[switch]{lineno}

\usepackage{stmaryrd}
\usepackage{enumitem}
\usepackage{dsfont}
\usepackage{adjustbox}
\usepackage{multirow}
\usepackage{xcolor}
\usepackage[table]{xcolor}
\usepackage{makecell}
\newcommand{\blue}[1]{\textcolor{blue}{#1}}
\newcommand{\Ctd}{\ensuremath{C^{\text{td}}}}
\newcommand{\ind}{\mathds{1}}
\newcommand{\bftab}{\fontseries{b}\selectfont}
\usepackage{bm}
\newcommand{\equal}[1]{{\hypersetup{linkcolor=black}\thanks{#1}}}

\theorembodyfont{\upshape}
\theoremheaderfont{\scshape}
\theorempostheader{:}
\theoremsep{\newline}

\jmlrvolume{297}
\jmlryear{2025}
\jmlrworkshop{Machine Learning for Health (ML4H) 2025} %

 \title[Deep Kernel Aalen-Johansen Estimator]{Deep Kernel Aalen-Johansen Estimator: An Interpretable and Flexible Neural Net Framework for Competing Risks}

 \author{%
  \Name{Xiaobin Shen}\equal{These authors contributed equally.} \Email{xiaobins@andrew.cmu.edu}\\
  \Name{George H. Chen}\footnotemark[1] \Email{georgechen@cmu.edu}\\
    \addr Carnegie Mellon University, Pittsburgh, PA, USA
 }

\begin{document}

\maketitle

\begin{abstract}
We propose an interpretable deep competing risks model called the Deep Kernel Aalen-Johansen (DKAJ) estimator, which generalizes the classical Aalen-Johansen nonparametric estimate of cumulative incidence functions (CIFs). Each data point (e.g., patient) is represented as a weighted combination of clusters. If a data point has nonzero weight only for one cluster, then its predicted CIFs correspond to those of the classical Aalen-Johansen estimator restricted to data points from that cluster. These weights come from an automatically learned kernel function that measures how similar any two data points are. On four standard competing risks datasets, we show that DKAJ is competitive with state-of-the-art baselines while being able to provide visualizations to assist model interpretation.
\end{abstract}

\paragraph*{Data and Code Availability}
We use standard publicly available competing risks benchmark datasets: PBC \citep{fleming1991counting}, Framingham \citep{kannel1979diabetes}, SEER\footnote{\url{https://seer.cancer.gov/}}, and the synthetic dataset by \citet{lee2018deephit}. Our code is available at: \mbox{\url{https://github.com/xiaobin-xs/dkaj}} %

\paragraph*{Institutional Review Board (IRB)}
This work does not require IRB approval as we analyze existing publicly available datasets.

\setlength{\abovedisplayskip}{2.5pt plus 1pt}
\setlength{\belowdisplayskip}{2.5pt plus 1pt}
\setlength{\abovedisplayshortskip}{1.5pt plus 1pt}
\setlength{\belowdisplayshortskip}{1.5pt plus 1pt}

\section{Introduction}
\label{sec:intro}

Various health applications involve reasoning about the amount of time that will elapse before a critical event happens, where there could be different types of critical events and we would like to know which type happens earliest and when. For example, for a patient in a hospital, we may want to model their time until either discharge or in-hospital mortality, where only one of these happens earliest. This is a well-studied problem referred to as the \emph{competing risks} setup in survival analysis literature (e.g., see Chapter~8 of the textbook by \citet{kalbfleisch2002statistical}).

Despite many recent methodological advances in developing flexible competing risks models, the bulk of the methods developed are not easy to interpret. For example, deep competing risks models like DeepHit \citep{lee2018deephit}, DeSurv \citep{danks2022derivative}, and Neural Fine-Gray \citep{jeanselme2023neural} are not designed to be straightforward to interpret. Similarly, gradient boosted trees for competing risks \citep{alberge2025survival} are also not easy to interpret as soon as there are many trees each with many leaves. Instead, for competing risks, if one wants to use a model that is inherently interpretable, the standard approach is still to use the classical Fine and Gray model \citep{fine1999proportional}, the cause-specific Cox model \citep{prentice1978analysis}, or slight variants of either of these.\footnote{Despite its name, the Neural Fine-Gray model \citep{jeanselme2023neural} does not closely resemble the original Fine and Gray model \citep{fine1999proportional} and is instead a competing risks extension of SuMo-net \citep{rindt2022survival}; i.e., Neural Fine-Gray does not inherit the interpretation advantages of the original Fine and Gray model.}

In this paper, we propose a new deep competing risks model that aims to be interpretable. We build upon the classical Aalen-Johansen (AJ) estimator \citep{aalen1978empirical}, which is a nonparametric approach developed for the competing risks setup that estimates population-level quantities. We show how to adapt the key ideas of the AJ estimator to instead produce predicted risks of different competing events at the individual data point level. What we get is a ``conditional'' AJ estimator, where we are conditioning on a test point's feature vector. A non-deep-learning-based conditional AJ estimator has recently been studied theoretically \citep{bladt2025conditional}. Our paper is the first that we are aware of that develops a deep-learning-based kernel AJ estimator.

Our deep kernel Aalen-Johansen (DKAJ) estimator represents each data point as a weighted combination of clusters (each cluster corresponds to a subset of training patients). Per cluster, we store summary information that corresponds to an AJ estimator restricted to data points (i.e., patients) in the cluster. Consequently, existing research on interpreting the AJ estimator can be used. Note that this interpretation is not ``post-hoc'' in that when making a prediction for a test point, if the test point is represented as a weighted combination that puts nonzero weight only on a single cluster, its prediction does correspond to the AJ estimator for that cluster. The weights in these weighted combinations are based on an automatically learned kernel function that measures how similar any two data points are.

From a technical viewpoint, the AJ estimator is the competing risks analogue of the classical Kaplan-Meier (KM) estimator \citep{kaplan1958nonparametric} that is meant for standard survival analysis without competing risks. Similar to how deep kernel KM estimators were developed by \citet{chen2020deep,chen2024survival}, our proposed method could be thought of as extending Chen's earlier deep kernel KM estimator approach to the competing risks setting, replacing the KM estimator with the AJ estimator. In fact, when there is only a single competing event type, our proposed method becomes the existing \emph{survival kernets} approach by \citet{chen2024survival} that does not handle competing risks.

\vspace{-1em}
\section{Background}
\label{sec:background}

We briefly review the statistical setup for survival analysis with competing risks (Section~\ref{sec:setup}) and the Aalen-Johansen (AJ) estimator (Section~\ref{sec:AJ}), both of which we present at a level of detail sufficient for understanding our deep kernel extension of the AJ estimator. Although our work generalizes the survival kernets method by \citet{chen2024survival}, we do not review it in this section and instead explain how it is a special case of our proposed method in Section~\ref{sec:DKAJ}.

\smallskip
\noindent
\textbf{Notation}~~
We use uppercase letters (e.g.,~$X$) to denote random variables  and lowercase letters (e.g.~$x$) to denote realizations of random variables as well as constants, although we also use uppercase $S$ to denote a survival function, as is standard in survival analysis literature. For any positive integer~$\ell$, we use the notation $[\ell]\triangleq\{1,2,\dots,\ell\}$. Meanwhile, we use $\ind\{\cdot\}$ to denote the indicator function that is~1 when its argument is true and 0 otherwise.

\vspace{-1em}
\subsection{Statistical Setup}
\label{sec:setup}
\vspace{-0.25em}

We use the setup in Section 6.1 of \citet{chen2024introduction}. We assume that there are $m$ critical event types that are mutually exhaustive, and that we have access to i.i.d.\ training data $(X_1,Y_1,\Delta_1),\dots,(X_n,Y_n,\Delta_n)$, where for the \mbox{$i$-th} subject, $X_i\in\mathcal{X}$ denotes the subject's feature vector\footnote{More generally, throughout the paper, the ``feature vector'' refers to time-independent covariates available at time~0 per data point. When non-tabular data are used (e.g., a medical image or ECG snippet), we assume that they are preprocessed into fixed-length embedding vectors that are treated as the ``feature vectors''. Dynamic prediction with time-varying covariates is beyond the scope of this paper and left for future work.}, $Y_i\in[0,\infty)$ denotes the time until the earliest critical event that happens or censoring, and $\Delta_i\in\{0,1,\dots,m\}$ is an event indicator stating which critical event type happens earliest (the special value of 0 means that censoring happens earliest). Each point $(X_i,Y_i,\Delta_i)$ is generated as follows:
\begin{enumerate}[leftmargin=*,itemsep=-4pt,topsep=1pt,partopsep=0pt]

\item Sample feature vector $X_i$ from $\mathbb{P}_{\mathsf{X}}$.

\item Sample the time $T_{i,\delta}$ until each event $\delta\in[m]$ happens. We do this jointly, i.e., we sample length-$m$ vector $\bm{T}_i\triangleq(T_{i,1},T_{i,2},\dots,T_{i,m})$ from distribution $\mathbb{P}_{\bm{\mathsf{T}}|\mathsf{X}}(\cdot|X_i)$, which we assume to be absolutely continuous (so ties among the $m$ entries in~$\bm{T}_i$ happen with probability~0). We denote the time of the earliest event as $T_i\triangleq\min_{\delta\in[m]}T_{i,\delta}$, and the earliest event that happens as $\Delta_i^*\triangleq\arg\min_{\delta\in[m]}T_{i,\delta}$.

\item Sample censoring time $C_i$ from absolutely continuous distribution $\mathbb{P}_{\mathsf{C}|\mathsf{X}}(\cdot|X_i)$.

\item Set $Y_i\triangleq\min\{T_i,C_i\}$, and
\[
\Delta_i\triangleq\begin{cases}
0 & \text{if }Y_i=C_i,\\
\Delta_i^* & \text{otherwise}.
\end{cases}
\]

\end{enumerate}
Steps 2 and 3 %
imply that conditioned on $X_i$, random variables $\bm{T}_i$ and $C_i$ are independent. %

\smallskip
\noindent
\textbf{Prediction task}~~
We assume that any test point is sampled using only the first two steps of the generative process for training data (i.e., we do not model a test point being censored). Namely:
\begin{enumerate}[leftmargin=*,itemsep=-4pt,topsep=1pt,partopsep=0pt]

\item Sample test feature vector $X$ from $\mathbb{P}_{\mathsf{X}}$.

\item Sample $\bm{T}\triangleq(T_1,\dots,T_m)$ from $\mathbb{P}_{\bm{\mathsf{T}}|\mathsf{X}}(\cdot|X)$, and set $T\triangleq\min_{\delta\in[m]}T_\delta$, and $\Delta^*\triangleq\arg\min_{\delta\in[m]}T_\delta$.

\end{enumerate}
We aim to predict the \emph{cumulative incidence function} (CIF), which is specific to each event type $\delta\in[m]$:
\begin{equation}
F_\delta(t|x)\triangleq\mathbb{P}(T\le t,\Delta^*=\delta\mid X=x)
\quad\text{for }t\ge0.
\label{eq:CIF}
\end{equation}
There are alternative ways to represent this CIF. For example, we can back out the CIF by knowing the \emph{event-specific hazard function} (also called the \emph{\mbox{cause-specific} hazard function} in the literature)
\begin{align*}
\lambda_\delta(t|x)
&\triangleq\lim_{h\downarrow0}\frac{\mathbb{P}\big(t\le T<t+h,\Delta^*=\delta~\!\big|~\!T\ge t,X=x\big)}{h}\\
&=\frac{f_\delta(t|x)}{S(t|x)},
\end{align*}
where
$
f_\delta(t|x)\triangleq\frac{\textrm{d}}{\textrm{d}t}F_\delta(t|x)
$
is the event-specific sub-density function, and
\begin{equation}
S(t|x)
\triangleq\mathbb{P}(T\!>\!t~\!|~\!X\!=\!x)
=\exp\!\bigg(
   -\sum_{\delta=1}^m
      \int_0^t\!\lambda_\delta(u|x)\textrm{d}u
 \bigg)
\label{eq:surv}
\end{equation}
is called the \emph{survival function}.
In particular, if we know the event-specific hazard function $\lambda_\delta(t|x)$ for all events $\delta\!\in\![m]$, then we can obtain $S(t|x)$ using equation~\eqref{eq:surv}, and subsequently recover $f_\delta(t|x) = \lambda_\delta(t|x)S(t|x)$. Finally, we can back out CIF $F_\delta(t|x)$ by integration: $F_\delta(t|x)=\int_0^t f_\delta(u|x)\textrm{d}u$.

\smallskip
\noindent
\textbf{Likelihood function}~~
The standard competing risks likelihood function that does not depend on the censoring distribution is given by
\begin{equation}
\mathcal{L}=\prod_{i=1}^{n}\big(\lambda_{\Delta_i}(Y_i|X_i)\big)^{\ind\{\Delta_i\ne0\}}S(Y_i|X_i).
\label{eq:likelihood}
\end{equation}
For details, see \citet[Section~8.2.3]{kalbfleisch2002statistical}. We rely on this likelihood function later as we use a maximum likelihood approach.

\vspace{-1em}
\subsection{Aalen-Johansen (AJ) Estimator}
\label{sec:AJ}
\vspace{-0.25em}

We now describe a simplified version of the AJ estimator (as stated by \citet{edwards2016methodologic} and implemented in the \texttt{lifelines} software package \citep{davidsonpilon2019lifelines}). %
For a more general introduction to the AJ estimator (for multistate processes), see, for instance, Chapter~3 of \citet{cook2018multistate}. The simplified AJ estimator we consider estimates the population-level CIF
$
F_\delta^{\text{pop}}(t) \triangleq {\mathbb{P}(T\le t, \Delta^*=\delta)}
$; this is ``population-level'' since it does not condition on $X$ as in equation~\eqref{eq:CIF}.

Let $0<t_1<t_2<\cdots<t_L$ denote the unique times in which any critical event happens in the training data (censoring is not a critical event). %
Next, we denote the number of critical events of type $\delta\in[m]$ that occur at time $t_\ell$ (with $\ell\in[L]$) by
\begin{equation}
d_{\delta,\ell}
\triangleq
  \sum_{j=1}^n \ind\{\Delta_j=\delta,Y_j=t_\ell\},
\label{eq:event-count}
\end{equation}
and the number of subjects ``at risk'' (who could possibly still experience any critical event) at time $t_\ell$ by
\begin{equation}
n_\ell
\triangleq
  \sum_{j=1}^n \ind\{Y_j\ge t_\ell\}.
\label{eq:at-risk-count}
\end{equation}
Then the simplified AJ estimator of $F_\delta^{\text{pop}}(t)$ is
\begin{equation}
\widehat{F}_\delta^{\text{AJ}}(t)
\triangleq \sum_{\ell=1}^L
  \frac{\ind\{t_\ell \le t\}d_{\delta,\ell}}
       {n_\ell}\widehat{S}^{\text{KM}}(t_{\ell-1})
\;\text{for }t\ge0,
\label{eq:simplified-AJ}
\end{equation}
where $t_0\triangleq0$, and $\widehat{S}^{\text{KM}}$ is the classical Kaplan-Meier estimator \citep{kaplan1958nonparametric}:\footnote{The Kaplan-Meier estimator is for the standard survival analysis setup without competing risks (equivalent to the competing risks setup with a single event type); here, all $m$ event types are lumped together into a single critical event type corresponding to any of the critical events happening.}
\begin{equation}
\widehat{S}^{\text{KM}}(t)
\triangleq
\prod_{\ell=1}^L
\Big(
  1 - 
  \frac{\sum_{\delta=1}^m d_{\delta,\ell}}
       {n_\ell}
\Big)^{\ind\{t_\ell \le t\}}
\;\text{for }t\ge0.
\label{eq:KM}
\end{equation}
The key observation we exploit in our derivation of our DKAJ estimator is that the simplified AJ estimator~\eqref{eq:simplified-AJ} can be derived by maximizing the likelihood $\mathcal{L}$ in equation~\eqref{eq:likelihood}, with some pre- and post-processing.
\begin{proposition}
\label{prop:AJ-MLE}
Using the notation above where $0<t_1<\cdots<t_L$ denote the unique times in which any critical event happens, suppose that we parameterize the event-specific hazard function to be piecewise constant on the $L$ intervals $(t_0,t_1], (t_1,t_2], \dots, (t_{L-1},t_L]$:
\begin{equation}
\lambda_\delta(t|x)
\!\triangleq\!
\begin{cases}
{\displaystyle \frac{\phi_{\delta,\ell}}{t_\ell-t_{\ell-1}}} & \!\!\text{if }t\!\in\!(t_{\ell-1},t_\ell]\text{ for }\ell\!\in\![L], \\
0 & \!\!\text{otherwise},
\end{cases}
\label{eq:piecewise-constant-hazard-pop-level}
\end{equation}
where $\phi_{\delta,\ell}\in[0,\infty)$ for $\delta\in[m], \ell\in[L]$ are the parameters. These parameters do not depend on~$x$, so the model here is at the population level. Thus, only for this section, we abbreviate $\lambda_\delta(t)\triangleq\lambda_\delta(t|x)$.

Suppose that we preprocess our training data $(X_1,Y_1,\Delta_1),\dots,(X_n,Y_n,\Delta_n)$ so that any point that is censored (so that $\Delta_i=0$) has its observed time $Y_i$ replaced by the latest time in which any critical event happened \textbf{prior to} $Y_i$ (if no critical event happened before $Y_i$, then set $Y_i=0$).\footnote{This preprocessing procedure is used by Breslow to derive the Breslow estimator \citep{breslow1972discussion} for the Cox proportional hazards model \citep{cox1972regression}.\label{foot:preprocessing}} %
After doing this preprocessing, the choice of parameters $\phi_{\delta,\ell}$ that maximizes the likelihood of equation~\eqref{eq:likelihood} is given by
\[
\widehat{\phi}_{\delta,\ell}
= \frac{d_{\delta,\ell}}{n_\ell}
\quad\text{for }\delta\in[m],\ell\in[L].
\]
Thus, the maximum likelihood estimator for the event-specific hazard function is---by plugging $\widehat{\phi}_{\delta,\ell}$ into $\phi_{\delta,\ell}$ in equation~\eqref{eq:piecewise-constant-hazard-pop-level}---given by
\begin{equation}
\!\!\widehat{\lambda}_\delta(t)
\!\triangleq\!
\begin{cases}
{\displaystyle \frac{d_{\delta,\ell}}{(t_\ell\!-\!t_{\ell-1})n_\ell}} & \!\!\text{if }t\!\in\!(t_{\ell-1},t_\ell]\text{ for }\ell\!\in\![L], \\
0 & \!\!\text{otherwise}.
\end{cases}
\label{eq:event-specific-hazard-MLE}
\end{equation}
The estimated CIF in equation~\eqref{eq:simplified-AJ} can be obtained via a form of interpolation based on this piecewise-constant maximum likelihood estimator of $\lambda_\delta(t|x)$. We defer details of this postprocessing step (interpolation) and the proof of this proposition to \appendixref{apd:prop-AJ-MLE}.
\end{proposition}

\vspace{-2.5em}
\section{Deep Kernel Aalen-Johansen}
\label{sec:DKAJ}
\vspace{-0.25em}

We now derive our neural net extension of the AJ estimator that uses a kernel function $K:\mathcal{X}\times\mathcal{X}\rightarrow[0,\infty)$ to measure how similar any two input points are (for two data points with feature vectors $x,x'\in\mathcal{X}$, their similarity score is given by $K(x,x')$, where higher means ``more similar''). We refer to our resulting approach as the \emph{deep kernel Aalen-Johansen} (DKAJ) estimator. %
This section is organized as follows:
\begin{itemize}[leftmargin=*,itemsep=-4pt,topsep=1pt,partopsep=0pt]
\item (Section~\ref{sec:neural-net-formulation}) We first explain how we parameterize the event-specific hazard function in terms of kernel function $K$ that is specified as a neural net, and how to maximize likelihood function~\eqref{eq:likelihood} to train neural net parameters. %
This training procedure amounts to learning an embedding space to represent data points (e.g., patients)~in.

\item (Section~\ref{sec:DKAJ-overall}) We then present our DKAJ model, which first runs the neural net training from Section~\ref{sec:neural-net-formulation}. Afterwards, we run exemplar-based clustering in the learned embedding space, and compute summary information per cluster. %
Any data point is then represented as a weighted combination of clusters. %
We explain why the simplified AJ estimator \eqref{eq:simplified-AJ} as well as survival kernets \citep{chen2024survival} are special cases of DKAJ, and we also discuss model interpretation.

\end{itemize}

\vspace{-1em}
\subsection{Neural Net Formulation and Loss}
\label{sec:neural-net-formulation}
\vspace{-.25em}

The key idea is that we modify the parameterization in equation~\eqref{eq:piecewise-constant-hazard-pop-level} to depend on feature vectors. Reusing notation from Section~\ref{sec:AJ} where $0<t_1<\cdots<t_L$ denote the unique times in which any critical event occurs in training data, we use the parameterization
\begin{align}
\lambda_\delta(t|x)
\!\triangleq\!
\begin{cases}
{\displaystyle \frac{\psi_{\delta,\ell}(x)}{t_\ell-t_{\ell-1}}} & \!\!\text{if }t\!\in\!(t_{\ell-1},t_\ell]\text{ for }\ell\!\in\![L], \\
0 & \!\!\text{otherwise},
\end{cases}
\label{eq:piecewise-constant-hazard}
\end{align}
where
\begingroup
\allowdisplaybreaks
\begin{align}
\psi_{\delta,\ell}(x)
&\triangleq
  \frac{d_{\delta,\ell}(x)}
       {n_\ell(x)}, \label{eq:psi} \\
d_{\delta,\ell}(x)
&\triangleq
  \sum_{j=1}^n \ind\{\Delta_j=\delta, Y_j=t_\ell\} K(x,X_j), \label{eq:kernel-event-count} \\
n_\ell(x)
&\triangleq
  \sum_{j=1}^n \ind\{Y_j\ge t_\ell\} K(x,X_j). \label{eq:kernel-at-risk-count}
\end{align}
\endgroup
Notice that $d_{\delta,\ell}(x)$ (equation~\eqref{eq:kernel-event-count}) and $n_\ell(x)$ (equation~\eqref{eq:kernel-at-risk-count}) generalize $d_{\delta,\ell}$ (equation~\eqref{eq:event-count}) and $n_\ell$ (equation~\eqref{eq:at-risk-count}), respectively. In particular, we use the kernel function to weight each training point $X_i$ by how similar it is to feature vector~$x$. If the kernel function were to always output~1, then we would just obtain the maximum likelihood estimate from equation~\eqref{eq:event-specific-hazard-MLE}.

Next, we parameterize the kernel function $K$ in the same manner as \citet{chen2020deep}:
\begin{equation}
K(x,x')\triangleq\exp(-\|f(x;\theta)-f(x';\theta)\|^2),
\label{eq:kernel-function}
\end{equation}
where $\|\cdot\|$ is Euclidean distance, and ${f(\cdot;\theta):\mathcal{X}\rightarrow\mathbb{R}^d}$ is a user-specified neural net such as a multilayer perceptron, with all neural net parameters collected into the variable~$\theta$ (standard strategies can be used such as choosing $f$ to be a CNN or vision transformer if ``feature vectors'' are instead images, etc).

At this point, combining equations~\eqref{eq:psi}, \eqref{eq:kernel-event-count}, \eqref{eq:kernel-at-risk-count}, and \eqref{eq:kernel-function}, we get
\begin{align}
&\psi_{\delta,\ell}(x;\theta) \nonumber\\
&\!\!=\!
  \frac{\sum_{j=1}^n \!\ind\{\Delta_j\!=\!\delta, Y_j\!=\!t_\ell\} \exp(-\|f(x;\theta)\!-\!f(X_j;\theta)\|^2)}
       {\sum_{j=1}^n \!\ind\{Y_j\!\ge\! t_\ell\} \exp(-\|f(x;\theta)\!-\!f(X_j;\theta)\|^2)},
\label{eq:psi-detailed}
\end{align}
where we now write ``$\psi_{\delta,\ell}(x;\theta)$'' instead of only ``$\psi_{\delta,\ell}(x)$'' to emphasize the dependence on $\theta$.

We state the log likelihood next with the help of equation~\eqref{eq:psi-detailed}. Let $\kappa(Y_i)\in\{0,1,\dots,L\}$ denote the time index corresponding to $Y_i$ (where we first apply the same preprocessing to censored times as mentioned in Proposition~\ref{prop:AJ-MLE}).
Then plugging in the event-specific hazard function $\lambda_\delta(t|x)$ into the log of the likelihood function $\mathcal{L}$ from equation~\eqref{eq:likelihood}, after a short derivation (for details, see \appendixref{apd:DKAJ-log-likelihood}), we obtain
\begin{align}
\!\!\!\!
\log\mathcal{L}(\theta)
&=\sum_{i=1}^n
    \sum_{\delta=1}^m
    \!
      \bigg[
        \ind\{\Delta_i\!=\!\delta\}
        \log\psi_{\delta,\kappa(Y_i)}(X_i;\theta) \nonumber\\
&\phantom{
   =\sum_{i=1}^n
     \sum_{\delta=1}^m\!
       \Bigg[~\!\!\!\!\!\!\!\!
 }
        -
        \sum_{\ell=1}^{\kappa(Y_i)}
          \psi_{\delta,\ell}(X_i;\theta)
      \bigg]\!+\!\text{constant}.\!
\label{eq:DKAJ-log-likelihood}
\end{align}
Maximizing this likelihood is problematic since the \mbox{$i$-th} training point's loss term uses a predicted hazard value that depends on the \mbox{$i$-th} training point's ground truth, causing overfitting. To fix this, we use the same strategy as \citet{chen2020deep} and instead aim to maximize the modified ``leave-one-out'' log likelihood
\begin{align}
\log\mathcal{L}^{\text{train}}(\theta)
&\triangleq
  \sum_{i=1}^n
    \sum_{\delta=1}^m\!
      \bigg[
        \ind\{\Delta_i\!=\!\delta\}
        \log\psi_{\delta,\kappa(Y_i)}^{\neg i}(X_i;\theta) \nonumber\\
&\phantom{
   =\sum_{i=1}^n
     \sum_{\delta=1}^m\!
       \Bigg[~
 }
        -
        \sum_{\ell=1}^{\kappa(Y_i)}
          \psi_{\delta,\ell}^{\neg i}(X_i;\theta)
      \bigg],
\label{eq:DKAJ-log-likelihood-train}
\end{align}
where
\begin{align*}
&\psi_{\delta,\ell}^{\neg i}(x;\theta)\\
&\!\!\triangleq\!
  \frac{\sum_{j\ne i} \!\ind\{\Delta_j\!=\!\delta, Y_j\!=\!t_\ell\} \exp(-\|f(x;\theta)\!-\!f(X_j;\theta)\|^2)}
       {\sum_{j\ne i} \!\ind\{Y_j\!\ge\! t_\ell\} \exp(-\|f(x;\theta)\!-\!f(X_j;\theta)\|^2)}.
\end{align*}
The numerator and denominator each sums over all training points except for the~$i$-th training point (i.e., $\psi_{\delta,\ell}^{\neg i}(x;\theta)$ is a leave-one-out version of $\psi_{\delta,\ell}(x;\theta)$). Note that equation~\eqref{eq:DKAJ-log-likelihood-train} also drops the constant term in equation~\eqref{eq:DKAJ-log-likelihood} as it does not affect the optimization.

In practice, we minimize the loss $-\frac{1}{n}\log\mathcal{L}^{\text{train}}(\theta)$ using minibatch gradient descent. In our experiments later, we further add a second loss term corresponding to DeepHit's ranking loss \citep{lee2018deephit}, so that the overall loss function we use is the same as DeepHit. As the details of this ranking loss is not essential to our exposition, we defer it to \appendixref{apd:deephit-loss}.

\vspace{3pt}
\noindent
\textbf{Neural net initialization}~~
We use the same scalable tree ensemble warm-start strategy as \citet{chen2024survival} referred to as Tree ensemble Under a Neural Approximation (TUNA), where the tree ensemble we use to warm-start is the recently developed competing risks model SurvivalBoost \citep{alberge2025survival}. We defer the details to \appendixref{apd:tuna-survivalboost}.

\vspace{-1em}
\subsection{Overall DKAJ Framework}
\label{sec:DKAJ-overall}
\vspace{-.5em}

The previous section shows how we can use minibatch gradient descent to learn neural net parameters stored in $\theta$. Effectively, once we learn these parameters, the neural net $f(\cdot;\theta)$ maps each data point to an embedding space. Following \citet{chen2024survival}, we now cluster in this embedding space and compute summary functions, 
where these are slightly different from Chen's since we are in a competing risks setup.

\vspace{3pt}
\noindent
\textbf{Training}~~
We train a DKAJ model as follows:
\begin{enumerate}[leftmargin=*,itemsep=-4pt,topsep=1pt,partopsep=0pt]

\item Use the training procedure in Section~\ref{sec:neural-net-formulation} to approximately solve $\widehat{\theta}\triangleq\arg\min_\theta -\frac{1}{n}\log\mathcal{L}^{\text{train}}(\theta)$. %

\item Compute embedding vector $\widetilde{X}_i\hspace{-2.75pt}\triangleq\hspace{-2.75pt}f(X_i;~\!\!\widehat{\theta})$ for ${i\!\in\![n]}$.

\item Run an exemplar-based clustering algorithm on embedding vectors $\widetilde{X}_1,\dots,\widetilde{X}_n$. For simplicity, we follow \citet{chen2024survival} and use $\varepsilon$-net clustering, which works as follows for a user-specified $\varepsilon>0$:

\begin{enumerate}[leftmargin=*,itemsep=-2pt,topsep=-2pt,partopsep=0pt]

\item Initialize the set of exemplars: $\mathcal{Q}\leftarrow\{1\}$ (training point~1 starts as the only exemplar).

\item For $i=2,\dots,n$:
\begin{enumerate}[leftmargin=*,itemsep=0pt,topsep=-2pt,partopsep=0pt]

\item Find the closest exemplar to $\widetilde{X}_i$: \\$j\leftarrow\arg\min_{q\in\mathcal{Q}}\|\widetilde{X}_i - \widetilde{X}_q\|$.

\item If $\|\widetilde{X}_i - \widetilde{X}_j\|\le\varepsilon$: assign training point~$i$ to be in exemplar $j$'s cluster.

Otherwise: set $\mathcal{Q}\rightarrow\mathcal{Q}\cup\{i\}$ \\ (make training point $i$ a new exemplar).

\end{enumerate}
\end{enumerate}
After clustering, let $\mathcal{C}_q\subseteq[n]$ denote the training points assigned to be in exemplar $q$'s cluster.

\item Per exemplar $q\in\mathcal{Q}$, compute the following summary functions:
\begin{align}
\!\!\!\!d_{\delta,\ell}^{\text{cluster}}(q)
&\triangleq
  \sum_{j\in\mathcal{C}_q} \ind\{\Delta_j=\delta, Y_j=t_\ell\} \nonumber\\[-.75em]
&\qquad\qquad~\!\text{for }\delta\in[m], \ell\in[L], \label{eq:cluster-event-count} \\[.25em]
\!\!\!\!n_\ell^{\text{cluster}}(q)
&\triangleq
  \sum_{j\in\mathcal{C}_q} \ind\{Y_j\ge t_\ell\}\quad\text{for }\ell\in[L]. \label{eq:cluster-at-risk-count}
\end{align}

\end{enumerate}
\noindent
\textbf{Prediction}~~
For any test feature vector $x$, prediction proceeds by using the following steps, where there is a hyperparameter $\tau>0$ for how close an exemplar has to be to the test point (in the embedding space) to contribute to the prediction for $x$:
\begin{enumerate}[leftmargin=*,itemsep=-4pt,topsep=1pt,partopsep=0pt]

\item Compute embedding vector $\widetilde{x}\rightarrow f(x;\widehat{\theta})$.

\item Find all exemplars in $\mathcal{Q}$ whose embedding vector is within Euclidean distance $\tau$ of $\widetilde{x}$. %
Denote the resulting set of close-enough exemplars as $\mathcal{Q}(x;\tau)$.

\item Compute weighted summary functions:
\begin{align}
d_{\delta,\ell}^{\text{DKAJ}}(x)
&\triangleq
  \sum_{q\in\mathcal{Q}(x;\tau)} K(x,X_q)d_{\delta,\ell}^{\text{cluster}}(q) \nonumber\\[-.25em]
&\qquad\qquad~\!\text{for }\delta\in[m], \ell\in[L], \label{eq:DKAJ-event-count} \\[.25em]
n_\ell^{\text{DKAJ}}(x)
&\triangleq
  \sum_{q\in\mathcal{Q}(x;\tau)} K(x,X_q)n_\ell^{\text{cluster}}(q) \nonumber\\[-.25em]
&\qquad\qquad~\!\text{for }\ell\in[L],\label{eq:DKAJ-at-risk-count}
\end{align}
where ${K(x,x')=\exp(-\|f(x;\widehat{\theta}) - f(x';\widehat{\theta})\|^2)}$.

\item Finally, we predict all events' CIFs. To do this, we first compute the survival function estimate
\[
\widehat{S}^{\text{DKAJ}}(t|x)
\triangleq \prod_{\ell=1}^L \Big( 1 - \frac{\sum_{\delta=1}^m d_{\delta,\ell}^{\text{DKAJ}}(x)}{n_{\ell}^{\text{DKAJ}}(x)} \Big)^{\ind\{t_\ell\le t\}}.
\]
Then we compute
\[
\!\!\!\!\!\!
\widehat{F}_\delta^{\text{DKAJ}}(t|x)
\triangleq
  \sum_{\ell=1}^L
    \frac{\ind\{t_\ell\le t\}d_{\delta,\ell}^{\text{DKAJ}}(x)}{n_\ell^{\text{DKAJ}}(x)}
    \widehat{S}^{\text{DKAJ}}(t_{\ell-1}|x).
\]
As a corner case, if the set of exemplars close enough to the test point is empty (i.e., ${\mathcal{Q}(x;\tau)=\emptyset}$), then simply predict the population-level Aalen-Johansen estimator from equation~\eqref{eq:simplified-AJ}.

\end{enumerate}
\noindent
\textbf{Special cases}~~
If $\varepsilon=0$ for $\varepsilon$-net clustering (so every training point is in its own cluster) and $\tau=\infty$ (so that there's effectively no distance threshold), then the predicted CIF simplifies to be of the form:
\[
\widehat{F}_\delta^{\text{DKAJ}}(t|x)
=\sum_{\ell=1}^L
   \frac{\ind\{t_\ell \le t\}d_{\delta,\ell}(x)}
        {n_\ell(x)}\widehat{S}^{\text{DKAJ}}(t_{\ell-1}|x),
\]
where
\[
\widehat{S}^{\text{DKAJ}}(t|x)
=
\prod_{\ell=1}^L
\Big(
  1 - 
  \frac{\sum_{\delta=1}^m d_{\delta,\ell}(x)}
       {n_\ell(x)}
\Big)^{\ind\{t_\ell \le t\}}.
\]
As a reminder, $d_{\delta,\ell}(x)$ and $n_\ell(x)$ are given in equations~\eqref{eq:kernel-event-count} and~\eqref{eq:kernel-at-risk-count}. When the training set size $n$ is large, this predicted CIF can be expensive to compute since, in principle, for any test point $x$, we would compute $K(x,X_i)$ for all $i\in[n]$. If furthermore $K(x,x')=1$ for all $x,x'$, then the DKAJ estimator becomes the population-level AJ estimator~\eqref{eq:simplified-AJ} (as we see shortly, another way to set DKAJ hyperparameters also yields the population-level AJ estimator).

As a second special case, if there is only one critical event type (i.e., $m=1$), then the overall DKAJ framework corresponds to survival kernets by \citet{chen2024survival}, where there would be no need to compute the CIF at the end (when there is only one critical event type, the survival function estimate $\widehat{S}^{\text{DKAJ}}(t|x)$ would be sufficient as the only predicted output).

As a third special case, we point out that in step~2 of the prediction procedure, if only one exemplar is within distance $\tau$ of $x$ in the embedding space (i.e., $|\mathcal{Q}(x;\tau)|=1$), then the predicted CIF corresponds to the AJ estimator from equation~\eqref{eq:simplified-AJ} applied to only data points in that single exemplar's cluster.

As an extreme example of the third special case, if for $\varepsilon$-net clustering, we set $\varepsilon=\infty$ (so all training points are in the same cluster), then the predicted CIF becomes the population-level AJ estimator~\eqref{eq:simplified-AJ}.

\smallskip
\noindent
\textbf{Model interpretation}~~
Any point with feature vector $x$ is represented as a weighted combination of clusters. In particular, from looking at equations~\eqref{eq:DKAJ-event-count} and~\eqref{eq:DKAJ-at-risk-count}, the only \mbox{exemplars} that contribute to the prediction for $x$ are the ones in the set $\mathcal{N}(x)\triangleq{\{q\in\mathcal{Q}(x;\tau):K(x,X_q)>0\}}$. Each exemplar $q$ in $\mathcal{N}(x)$ has weight $K(x,X_q)$; in fact we can normalize the weights to sum to 1.\footnote{For each $x$, we can normalize the weights $K(x,X_q)$ for $q\in\mathcal{N}(x)$ to sum to~1 since the normalization factor would cancel out when we divide equation~\eqref{eq:DKAJ-event-count} by~\eqref{eq:DKAJ-at-risk-count} in the calculation of $\widehat{F}_\delta^{\text{DKAJ}}(t|x)$ as well as $\widehat{S}^{\text{DKAJ}}(t|x)$).} Thus, we know precisely which exemplars/clusters contribute to the prediction for $x$ and with what weights.

From our discussion of special cases above, a key takeaway is that when using $\varepsilon$-net clustering, as $\varepsilon\rightarrow\infty$, the number of clusters decreases until there is a single cluster (and when $\varepsilon=0$, every training point is its own cluster). A DKAJ model is easier to interpret if it has fewer, coarser clusters (corresponding to when $\varepsilon$ is larger). In practice, one could tune $\varepsilon$ based on an evaluation metric on the validation set. %

After training a DKAJ model, it is possible to make visualizations to interpret any cluster as well as any data point. We defer showing these visualizations to the next section when we cover experiments (our visualizations depend on our experimental setup).

\smallskip
\noindent
\textbf{Optional summary function fine-tuning}~~
Survival kernets \citep{chen2024survival} applies an optional final training step that treats $d_{\delta,\ell}^{\text{cluster}}(q)$ and $n_\ell^{\text{cluster}}(q)$ in equations~\eqref{eq:cluster-event-count} and~\eqref{eq:cluster-at-risk-count} as trainable neural nets. %
We can fine-tune these summary functions to minimize the negative log likelihood loss. %
The prediction procedure remains the same. Extending Chen's strategy to handle competing risks is straightforward; for details, see \appendixref{apd:SFT}. For simplicity, in the main paper we do not show results using this optional step but we show results with this optional step in \appendixref{apd:exp-results}. %

\vspace{-1em}
\section{Experiments}
\label{sec:experiments}
\vspace{-.25em}

We now turn to numerical experiments, where our main goals are to show how well DKAJ works in practice (benchmarked against classical and state-of-the-art baselines on standard datasets) and to also illustrate how to make visualizations to interpret a trained DKAJ model. %

\begin{table*}[hbtp]
    \centering \setlength{\tabcolsep}{5pt}
    \floatconts {tab:dataset-summary}
    {\caption{Dataset summary statistics. Every dataset has two competing event types (primary vs competing).}\vspace{-2em}}
    {\setlength{\tabcolsep}{3pt}
    \begin{adjustbox}{max width=0.93\textwidth}
    \begin{tabular}{cccccc}
    \toprule
        \bfseries Dataset & \bfseries Number of Data Points & \bfseries Features & \bfseries Primary Event & \bfseries Competing Event & \bfseries Censoring Rate \\ \midrule
        PBC & 1,945 & 15 & Death (37.3\%) & Transplant (7.6\%) & 55.2\% \\ 
        Framingham & 4,434 & 18 & CVD Death (26.1\%) & Non-CVD Death (17.7\%) & 56.2\% \\ 
        SEER & 24,907 & 22 & BC (16.7\%) & CVD (4.4\%) & 78.9\% \\ 
        Synthetic & 30,000 & 12 & Event~1 (25.3\%) & Event~2 (24.7\%) & 50.0\% \\ 
    \bottomrule
    \end{tabular}
    \end{adjustbox}\vspace{-1.2em}
    }
\end{table*}

\vspace{3pt}
\noindent
\textbf{Datasets}~~
We use the following four standard competing risks datasets (see \tableref{tab:dataset-summary} for summary statistics), each with two critical event types (we refer to one as ``primary'' and the other as ``competing''):
\begin{itemize}[leftmargin=*,itemsep=-4pt,topsep=1pt,partopsep=0pt]
    \item PBC \citep{fleming1991counting}: a cohort of 1,945 patients with primary biliary cirrhosis (PBC) described by 15 features. The primary event is death, and the competing event is liver transplant.
    \item Framingham \citep{kannel1979diabetes}: a %
    longitudinal study of 4,434 participants with 18 baseline features aimed at identifying cardiovascular disease (CVD) risk factors. We restrict our analysis to the earliest available feature measurements per patient to make the dataset tabular and not longitudinal. The primary event is death from CVD, with death from other causes as the competing event.
    \item SEER: a subset of 24,907 female breast cancer patients from the SEER\footnote{\url{https://seer.cancer.gov/}} registry, restricted to diagnoses in the year 2010, and described by 22 demographic and disease-related features. The primary event is breast cancer (BC) mortality, with cardiovascular disease (CVD) as the competing event.
    \item Synthetic \citep{lee2018deephit}: a simulated dataset of 30,000 individuals with 12 features, constructed with two hypothetical competing events and subject to 50\% random censoring.
\end{itemize}
\vspace{3pt}
\noindent
\textbf{Baselines}~~
We use the following baselines:
\begin{itemize}[leftmargin=*,itemsep=-4pt,topsep=1pt,partopsep=0pt]
\item Classical competing risks models: the Fine and Gray model (abbreviated as Fine-Gray) \citep{fine1999proportional}, cause-specific Cox (cs-Cox) \citep{prentice1978analysis}, and random survival forest with competing risks (RSF-CR) \citep{ishwaran2014random}.
\item Deep learning–based methods: %
DeepHit \citep{lee2018deephit}, Deep Survival Machines (DSM) \citep{nagpal2021deep}, and Neural Fine-Gray (NeuralFG) \citep{jeanselme2023neural}.
\item SurvivalBoost \citep{alberge2025survival}, a recently developed gradient boosted tree ensemble model for competing risks.
\end{itemize}
Aside from Fine-Gray and cs-Cox, the other baselines are not designed to be inherently interpretable.\footnote{RSF-CR is typically used with a large number of trees each with many leaves so that interpretation is not straightforward. Separately, when we talk about a model being ``inherently interpretable'', we mean that the model's prediction itself comes with an explanation that does not rely on general post-hoc explanation tools such as Shapley-value-based approaches (e.g., \citealt{shapley1953value,lundberg2017unified}) or feature importance by permutation (e.g., \citealt{breiman2001random,ishwaran2014random,fisher2019all}).}

\vspace{3pt}
\noindent
\textbf{Evaluation metrics}~~
Model performance is assessed using time-dependent concordance index ($C^{\text{td}}$) \citep{antolini2005time} and integrated Brier score (IBS) for competing risks. Formal definitions and implementation details are provided in \appendixref{apd:eval-metric}.

\vspace{3pt}
\noindent
\textbf{Experimental setup}~~
For each dataset, we randomly split the data into 70\% training and 30\% testing subsets. Within the training subset, we further split the data into 80\% ``proper'' training and 20\% validation subsets. Models are trained on the proper training set under varying hyperparameter configurations, and the validation set is used for model selection. Final performance is reported on the held-out test set. For every dataset and method, this procedure is repeated 10 times with different random splits. Details of the hyperparameter search space and optimization strategy are provided in \appendixref{apd:exp-hyperparam}.

\begin{table}[!t]
    \centering
    \floatconts {tab:main-result-ctd}
    {\caption{\vspace{-.75em}Test set $C^{\text{td}}$ (mean$~\!\pm~\!$std.~dev.~across 10 random splits) for primary and competing events (best score in \textbf{bold}, 2nd best in \blue{blue}).}\vspace{-1.8em}}
    {
    \begin{adjustbox}{max width=0.45\textwidth}
    \begin{tabular}{cccc}
        \toprule
        \bfseries Dataset & \bfseries Method & \bfseries Primary & \bfseries Competing \\ 
        \midrule
        \multirow{8}{*}{\rotatebox{90}{PBC}} & Fine-Gray & 0.8212$\pm$0.0119 & 0.8769$\pm$0.0203 \\
        ~ & cs-Cox & 0.8295$\pm$0.0102 & 0.9094$\pm$0.0132 \\
        ~ & RSF-CR & \blue{0.8584$\pm$0.0089} & 0.8746$\pm$0.0163 \\
        ~ & DeepHit & 0.8407$\pm$0.0131 & 0.9060$\pm$0.0130 \\
        ~ & DSM & 0.8319$\pm$0.0115 & 0.9039$\pm$0.0189 \\
        ~ & NeuralFG & 0.8363$\pm$0.0150 & \blue{0.9120$\pm$0.0149} \\
        ~ & SurvivalBoost & \bftab{0.8719$\pm$0.0107} & \bftab{0.9360$\pm$0.0114} \\
        ~ & DKAJ & 0.8413$\pm$0.0105 & 0.9039$\pm$0.0170 \\
        \midrule
        \multirow{8}{*}{\rotatebox{90}{Framingham}} & Fine-Gray & \blue{0.7733$\pm$0.0106} & \blue{0.7144$\pm$0.0182} \\
        ~ & cs-Cox & \bftab{0.7753$\pm$0.0105} & \bftab{0.7160$\pm$0.0173} \\
        ~ & RSF-CR & 0.7720$\pm$0.0098 & 0.6982$\pm$0.0119 \\
        ~ & DeepHit & 0.7423$\pm$0.0198 & 0.6957$\pm$0.0195 \\
        ~ & DSM & 0.7664$\pm$0.0170 & 0.7095$\pm$0.0184 \\
        ~ & NeuralFG & 0.6833$\pm$0.0619 & 0.6978$\pm$0.0287 \\
        ~ & SurvivalBoost & 0.7645$\pm$0.0152 & 0.7031$\pm$0.0156 \\
        ~ & DKAJ & 0.7677$\pm$0.0138 & 0.7076$\pm$0.0209 \\
        \midrule
        \multirow{8}{*}{\rotatebox{90}{SEER}} & Fine-Gray & 0.8151$\pm$0.0021 & 0.8478$\pm$0.0099 \\
        ~ & cs-Cox & 0.7630$\pm$0.0085 & 0.8415$\pm$0.0132 \\
        ~ & RSF-CR & \blue{0.8350$\pm$0.0030} & 0.8301$\pm$0.0137 \\
        ~ & DeepHit & 0.8239$\pm$0.0028 & 0.8548$\pm$0.0120 \\
        ~ & DSM & 0.7807$\pm$0.0088 & 0.8422$\pm$0.0119 \\
        ~ & NeuralFG & 0.7743$\pm$0.0090 & 0.8362$\pm$0.0138 \\
        ~ & SurvivalBoost & \bftab{0.8418$\pm$0.0038} & \blue{0.8583$\pm$0.0081} \\
        ~ & DKAJ & 0.8333$\pm$0.0034 & \bftab{0.8598$\pm$0.0072} \\
        \midrule
        \multirow{8}{*}{\rotatebox{90}{Synthetic}} & Fine-Gray & 0.5823$\pm$0.0051 & 0.5917$\pm$0.0071 \\
        ~ & cs-Cox & 0.5809$\pm$0.0051 & 0.5902$\pm$0.0073 \\
        ~ & RSF-CR & 0.7224$\pm$0.0071 & 0.7203$\pm$0.0043 \\
        ~ & DeepHit & \blue{0.7401$\pm$0.0067} & \blue{0.7437$\pm$0.0043} \\
        ~ & DSM & 0.7280$\pm$0.0055 & 0.7320$\pm$0.0042 \\
        ~ & NeuralFG & \bftab{0.7481$\pm$0.0073} & \bftab{0.7513$\pm$0.0045} \\
        ~ & SurvivalBoost & 0.7160$\pm$0.0083 & 0.7190$\pm$0.0039 \\
        ~ & DKAJ & 0.7384$\pm$0.0067 & 0.7435$\pm$0.0041 \\
        \bottomrule
    \end{tabular}
    \vspace{-2em}
    \end{adjustbox}
    }
\end{table}

\vspace{3pt}
\noindent
\textbf{Model performance}~~
Table~\ref{tab:main-result-ctd} reports the test set $C^{\text{td}}$ values (mean $\pm$ standard deviation) across 10 different experimental repeats. For each run, hyperparameters are selected based on the model achieving the best validation $C^{\text{td}}$.
We also performed model selection using the validation IBS as the criterion. The corresponding results for both $C^{\text{td}}$ and IBS are provided in \appendixref{apd:exp-results} under both circumstances.\footnote{The choice of validation criterion has a substantial impact: selecting models based on IBS generally yields lower $C^{\text{td}}$, and conversely, selecting based on $C^{\text{td}}$ tends to yield worse IBS. This issue is well-known in survival analysis literature (e.g., \citealt{lillelund2025stop}), which motivates our decision to report both metrics under both model selection strategies.\label{foot:validation-criterion}} Model performance with the optional summary function fine-tuning is also reported in \appendixref{apd:exp-results}.

We highlight two key takeaways: first, no single method consistently outperforms the others across all datasets and event types. Second, our proposed DKAJ estimator achieves performance that is competitive with state-of-the-art baselines. In more detail, we conduct paired Wilcoxon signed-rank tests across all methods (\appendixref{apd:pairwise-tests}), and find that DKAJ achieves significantly lower IBS and competitive or higher \Ctd~relative to most baselines, while differences among the top-performing models are generally not significant. %
DKAJ is also amenable to model interpretation, which we explore momentarily in terms of cluster-level and individual-level visualizations. Again, the only baselines we compared against that are interpretable are Fine-Gray and \mbox{cs-Cox}, both of which provide a population-level notion of interpretability (e.g., global hazard ratios).

\vspace{3pt}
\noindent
\textbf{Ablation study}~~
We remove two components of DKAJ to assess their impact on model performance: (1) the leave-one-out (LOO) likelihood in equation~\eqref{eq:DKAJ-log-likelihood-train}, replaced by the standard likelihood in equation~\eqref{eq:DKAJ-log-likelihood}; 
and (2) the TUNA warm start, replaced by PyTorch’s default initialization. 
As shown in Table~\ref{tab:ablation-results}, for all three real-world datasets (PBC, Framingham, and SEER), removing either component results in worse performance in most cases.
Details are provided in Appendix~\ref{apd:ablation}.

\vspace{3pt}
\noindent
\textbf{Model scalability}~~
We assess model scalability in two manners. First, we vary the training set size from 10\% to 70\% on the synthetic dataset, with the remaining 30\% held out for testing (see Figure~\ref{fig:C-td-varying-training-size-e1} for results on Event 1). 
Across all models, the $C^{\text{td}}$ improves with larger training sets and stabilizes once sufficient data are available. DKAJ exhibits consistent gains without signs of overfitting or instability, maintaining performance comparable to other neural and ensemble baselines across all data regimes. Results for Event 2 show similar trends (Appendix~\ref{apd:varying-training-size}). We separately show that DKAJ has computation cost (wall-clock training time and model size) comparable to recently developed methods (Appendix~\ref{apd:compute-cost}).

\begin{figure}
\floatconts
  {fig:C-td-varying-training-size-e1}
  {\caption{\Ctd~on the synthetic dataset (Event 1) as training set size varies; 30\% held-out test}\vspace{-1.5em}}
  {\includegraphics[width=.95\linewidth]{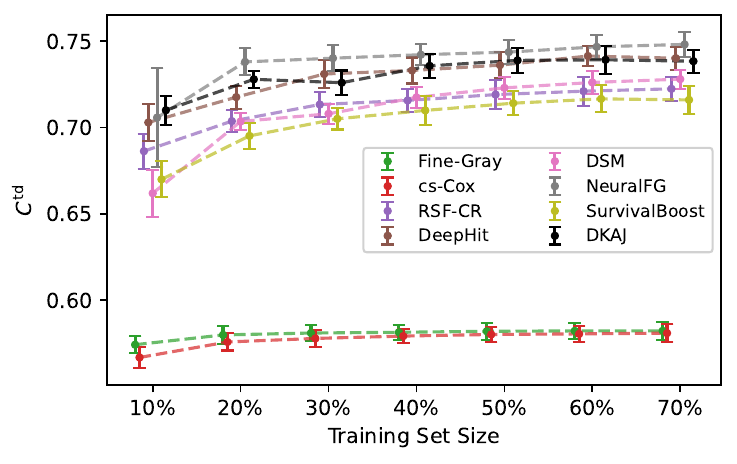}\vspace{-2.7em}}
\end{figure}

\vspace{3pt}
\noindent
\textbf{Cluster-level visualizations}~~
For a trained DKAJ model, we can visualize the (a) CIFs of any cluster, and also (b) a heatmap showing the distributions of features of any cluster. We illustrate these two visualizations (a) and (b) in Figures~\ref{fig:clusters-aj-top5-framingham} and \ref{fig:clusters-heatmap-top5-framingham} respectively for the Framingham dataset, where we restrict our attention to the five largest clusters for simplicity (largest meaning the most number of training points assigned to them). We then sort these clusters 
by the estimated probability of experiencing the primary event (CVD death) earliest. Specifically, clusters can be ranked by the CIF of the primary event evaluated at the maximum observed training event time~$t_{\max}$.  %

\begin{figure}[t!]
\floatconts
  {fig:clusters-aj-top5-framingham}
  {\caption{(Framingham) CIFs for the largest 5 clusters (we then sort these 5 clusters in decreasing order by the estimated probability of CVD death happening within the maximum observed time). Clusters correspond across the two plots in this figure as well as in Figure~\ref{fig:clusters-heatmap-top5-framingham} (e.g., the blue cluster is the same cluster across these visualizations).}\vspace{-2.5em}} %
  {\vspace{-.5em}\includegraphics[width=.95\linewidth]{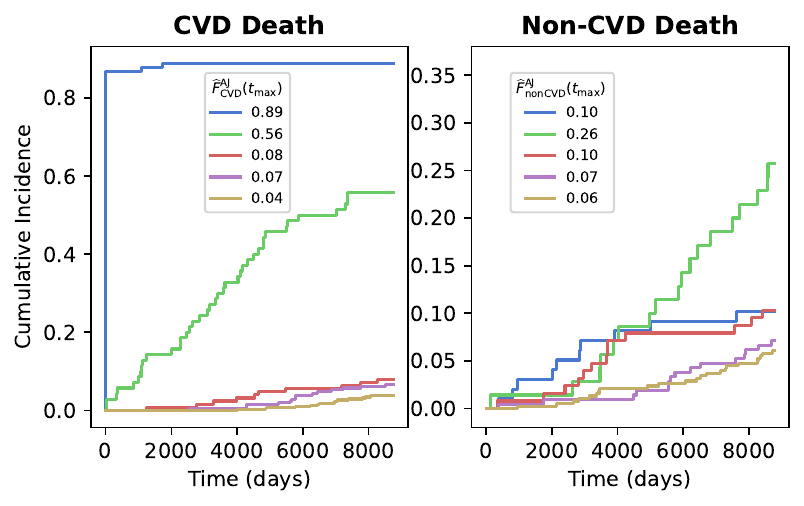}\vspace{-2.5em}}
\end{figure}

In more detail, Figure~\ref{fig:clusters-aj-top5-framingham} shows the CIFs for the 5 largest clusters, ordered by the risk of CVD death (i.e., CIF of CVD death evaluated at $t_{\max}$); this risk is also displayed per cluster in the legend. The corresponding clusters' feature distributions are shown in Figure~\ref{fig:clusters-heatmap-top5-framingham}. For example, the blue cluster (estimated risk of CVD death $=$ 0.89) exhibits high CVD risk and consists of patients with a history of hypertension (\texttt{PREVHYP}), coronary heart disease (\texttt{PREVCHD}), angina pectoris (\texttt{PREVAP}), current smoking (\texttt{CURSMOKE}), and older age. The green cluster (estimated risk of CVD death $=$ 0.56) also shows high CVD mortality, but in addition displays an elevated risk of death from other causes; this cluster is characterized by higher blood pressure (\texttt{SYSBP}, \texttt{DIABP}) and BMI, but a lower prevalence of prior coronary heart disease, suggesting a different clinical profile. In contrast, the other three clusters have substantially lower cumulative incidence of both CVD and non-CVD death. Such visualizations highlight similarities and differences between high-risk groups, and illustrate how the model's clusters align with meaningful clinical subpopulations.

A kernel-similarity heatmap for Framingham (50 largest clusters) is provided in Appendix~\ref{apd:visualization}, Figure~\ref{fig:kernel-matrix-top50-clust-framingham}, showing a clear block-diagonal pattern consistent with coherent clusters of varying sizes.

Visualizations for the 5 largest clusters in %
PBC and SEER datasets are provided in Appendix~\ref{apd:visualization}. We omit visualizations for the synthetic dataset as this dataset does not have a clinical interpretation to begin with.

\begin{figure}[t!]
\floatconts
  {fig:clusters-heatmap-top5-framingham}
  {\caption{(Framingham) Feature heatmap summarizing distributions of variables in the same 5 clusters as in Figure~\ref{fig:clusters-aj-top5-framingham}. Darker shades mean higher feature values or frequencies.\vspace{-2.6em}}}
  {\includegraphics[width=.95\linewidth]{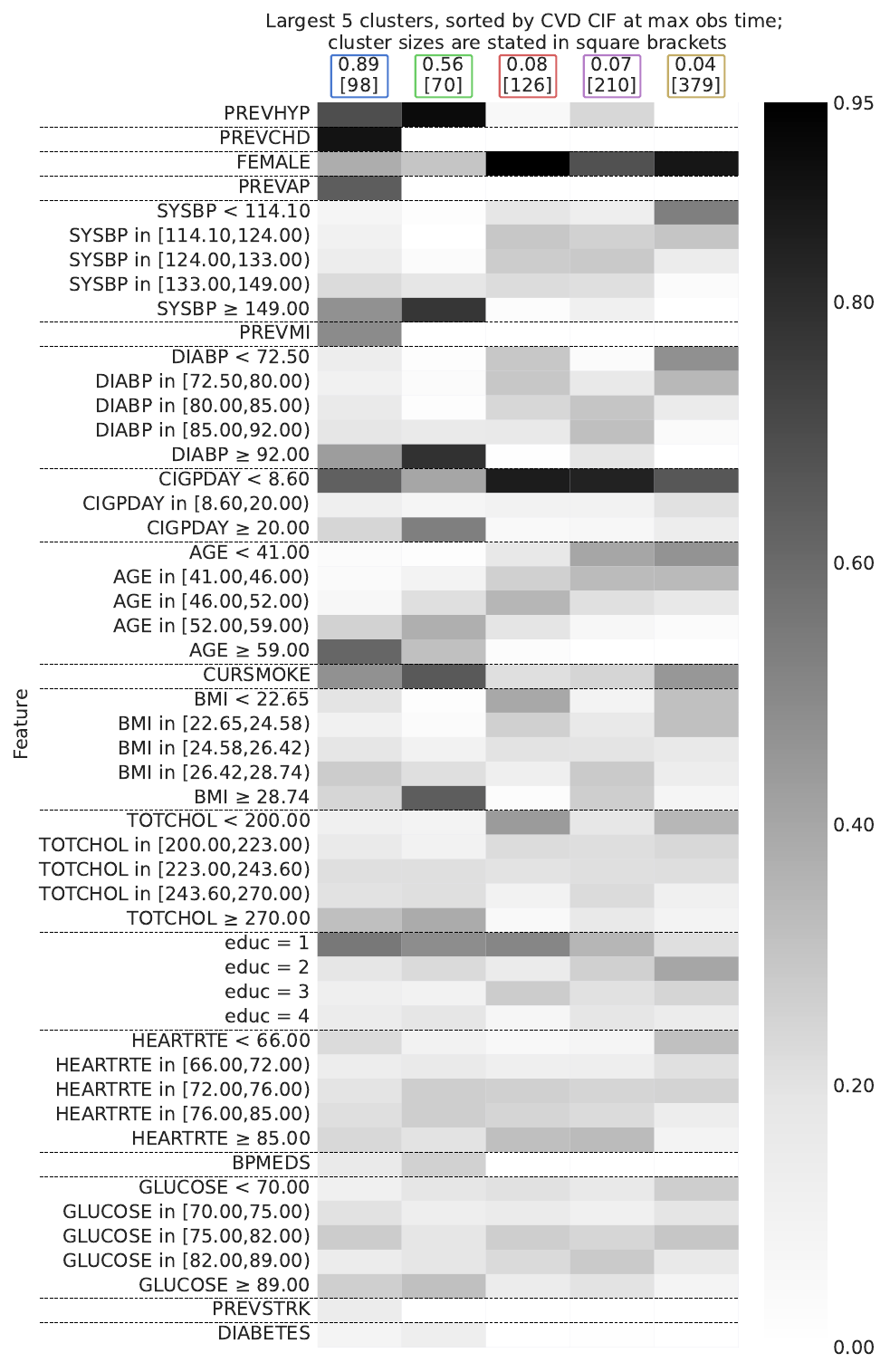}\vspace{-2.5em}}
\end{figure}

\vspace{2pt}
\noindent
\textbf{Individual-level visualizations}~~
Per test subject, we can examine the subject’s predicted CIF curves directly. In addition, the cluster weights assigned by the DKAJ model provide an interpretable decomposition: by identifying which clusters contribute most strongly to a prediction, we can inspect the defining characteristics of those clusters and thus better understand which clinical factors the model considers most relevant for that individual. Furthermore, conditional median times to each event, given that event occurs earliest, can be reported. Examples of individual-level visualizations are in Appendix~\ref{apd:visualization}.

\vspace{-.25em}
\vspace{-.75em}
\section{Discussion}
\label{sec:discussion}
\vspace{-.25em}

We have proposed a new flexible and interpretable competing risks model DKAJ that is competitive with various state-of-the-art baselines. %
Some possible extensions include using a proper scoring rule (such as the one by \citet{alberge2025survival}) as a training loss, figuring out better ways to tune hyperparameters especially to account for ease of model interpretation, trying other competing risks evaluation metrics (e.g., PDI \citep{ding2021evaluation}), and generalizing beyond competing risks to multistate processes.

We would like to provide some commentary regarding DKAJ's kernel function formulation and choice of clustering method. For simplicity, we defined the kernel function to resemble a Gaussian kernel but other options are possible (e.g., Laplacian, Cauchy). Moreover, it is possible to have the kernel function be time-dependent (this is done by SurvivalBoost; see Appendix~\ref{apd:tuna-survivalboost}), or to learn a separate kernel function per critical event type---although these make model interpretation more involved. For clustering, we follow \citet{chen2024survival} and use $\varepsilon$-net clustering. Chen used $\varepsilon$-net clustering as it enables establishing a theoretical error bound for survival kernets. We suspect that establishing a similar such guarantee for DKAJ is also possible but defer this to future work.
In practice, other exemplar-based clustering methods (e.g., affinity propagation, k-medoids) can also be used. Technically, using a non-exemplar-based clustering method like k-means can be done but the cluster centers then would no longer correspond to actual training patients, which can complicate model interpretation. %

\clearpage

\acks{This work was supported by NSF CAREER award \#2047981. The authors thank Yu Cheng, Ying Ding, and the anonymous reviewers for helpful feedback.}

\bibliography{references}

\clearpage
\appendix

\renewcommand{\thefigure}{\thesection.\arabic{figure}}
\renewcommand{\thetable}{\thesection.\arabic{table}}

\counterwithin{figure}{section}
\counterwithin{table}{section}

\section{Details on Proposition~\ref{prop:AJ-MLE}: Proof with Explanation of Interpolation}
\label{apd:prop-AJ-MLE}

Let $0<t_1<t_2<\cdots<t_L$ denote the unique times in which any critical event occurs in the training data. We also define $t_0\triangleq0$. Collectively, $t_0<t_1<\cdots<t_L$ forms our discrete time grid. For an observed time $Y_i$ that corresponds to an uncensored data point (i.e., $\Delta_i\ne0$), let $\kappa(Y_i)\in[L]$ denote the time index corresponding to $Y_i$. Following the preprocessing that we referenced in footnote \ref{foot:preprocessing}, we take a censored data point's observed time to be the preceding
uncensored event time. Thus, if $\Delta_i=0$, then
\[
\kappa(Y_i)=\max\big\{\ell\in\{0,1,\dots,L\}\text{ s.t.\ }t_\ell<Y_i\}.
\]
We use the parameterization from equation~\eqref{eq:piecewise-constant-hazard-pop-level}, which we reproduce below for convenience:
\begin{align*}
\lambda_\delta(t|x)
=\begin{cases}
{\displaystyle \frac{\phi_{\delta,\ell}}{t_\ell-t_{\ell-1}}} & \text{if }t\in(t_{\ell-1},t_\ell]\text{ for }\ell\in[L],\\
0 & \text{otherwise}.
\end{cases}
\tag*{(equation~\eqref{eq:piecewise-constant-hazard-pop-level}, reproduced)}
\end{align*}
As a reminder, this parameterization does not depend on the input feature vector~$x$, so that we can write $\lambda_\delta(t|x)\triangleq\lambda_\delta(t)$. We plug this parameterization into the log of the likelihood $\mathcal{L}$ in equation~\eqref{eq:likelihood} to get
\begin{align}
&\log\mathcal{L} \nonumber\\
&=\sum_{\delta=1}^m
    \sum_{i=1}^n\!
      \Bigg[
        \ind\{\Delta_i\!=\!\delta\}\log \lambda_\delta(Y_i|X_i)
        \!-\!
        \int_0^{Y_i}\!\lambda_\delta(u|X_i)\textrm{d}u
      \Bigg] \nonumber\\
&=\sum_{\delta=1}^m
    \sum_{i=1}^n\!
      \Bigg[
        \ind\{\Delta_i\!=\!\delta\}\log\phi_{\delta,\kappa(Y_i)} \nonumber\\
&\phantom{
   =\sum_{\delta=1}^m
      \sum_{i=1}^n\!
        \Bigg[
 }
        -
        \sum_{\ell=1}^{\kappa(Y_i)}
          \frac{\phi_{\delta,\ell}}{t_\ell-t_{\ell-1}}(t_\ell-t_{\ell-1})
      \Bigg] \nonumber\\
&=\sum_{\delta=1}^m
    \sum_{i=1}^n\!
      \Bigg[
        \ind\{\Delta_i\!=\!\delta\}\log\phi_{\delta,\kappa(Y_i)}
        -
        \sum_{\ell=1}^{\kappa(Y_i)}
          \phi_{\delta,\ell}
      \Bigg] \nonumber\\
&=\sum_{\delta=1}^m
    \sum_{i=1}^n
      \sum_{\ell=1}^{\kappa(Y_i)}\!
        \big[
          \ind\{\kappa(Y_i)\!=\!\ell,\Delta_i\!=\!\delta\}
          \log\phi_{\delta,\ell}
          -
          \phi_{\delta,\ell}
        \big] \nonumber\\
&=\sum_{\delta=1}^m
    \sum_{\ell=1}^L
      \sum_{i=1}^n
        \big[
          \ind\{\kappa(Y_i)\!=\!\ell,\Delta_i=\delta\}
          \log\phi_{\delta,\ell} \nonumber\\[-1em]
&\phantom{
   =\sum_{\delta=1}^m
      \sum_{\ell=1}^L
        \sum_{i=1}^n
          \big[
 }
          -
          \ind\{\kappa(Y_i)\!\ge\!\ell\}\phi_{\delta,\ell}
        \big].
\label{eq:AJ-log-likelihoood-calculation}
\end{align}
Note that the condition $\kappa(Y_i)=\ell$ is equivalent to $Y_i=t_\ell$. Similarly, the condition $\kappa(Y_i)\ge\ell$ is equivalent to $Y_i\ge t_\ell$. Thus,
\begin{align*}
&\log\mathcal{L}
 =\sum_{\delta=1}^m
    \sum_{\ell=1}^L
      \sum_{i=1}^n
        \big[
          \ind\{Y_i\!=\!t_\ell,\Delta_i=\delta\}
          \log\phi_{\delta,\ell}\\[-1em]
&\phantom{
   \log\mathcal{L}
   =\sum_{\delta=1}^m
      \sum_{\ell=1}^L
        \sum_{i=1}^n
          \big[
 }
          -
          \ind\{Y_i\!\ge\!t_\ell\}\phi_{\delta,\ell}
        \big].
\end{align*}
Setting the derivative with respect to $\phi_{\delta,\ell}$ to 0, we get
\begin{align*}
0
&=\Big[
    \frac{\partial\log\mathcal{L}}
         {\partial\phi_{\delta,\ell}}
  \Big]_{\phi_{\delta,\ell}=\widehat{\phi}_{\delta,\ell}} \\
&=\frac{\sum_{i=1}^n
          \ind\{Y_i\!=\!t_\ell,\Delta_i\!=\!\delta\}}
       {\widehat{\phi}_{\delta,\ell}}
  -
  \sum_{i=1}^n
    \ind\{Y_i\ge t_\ell\}.
\end{align*}
Rearranging terms yields
\[
\widehat{\phi}_{\delta,\ell}
=\frac{\sum_{i=1}^n \ind\{Y_i\!=\!t_\ell,\Delta_i\!=\!\delta\}}
      {\sum_{i=1}^n \ind\{Y_i\!\ge t_\ell\}}
=\frac{d_{\delta,\ell}}{n_\ell}.
\]
One can verify that $[\frac{\partial^2\log\mathcal{L}}{\partial\phi_{\delta,\ell}^2}]_{\phi_{\delta,\ell}=\widehat{\phi}_{\delta,\ell}}<0$ so that $\widehat{\phi}_{\delta,\ell}$ is a maximum (in fact, it's the global maximum since $\log\mathcal{L}$ is concave). Moreover, $\widehat{\phi}_{\delta,\ell}$ is clearly nonnegative (although we did not explicitly enforce this constraint in the optimization, it is a constraint that does need to be satisfied). Plugging in $\widehat{\phi}_{\delta,\ell}$ in place of $\phi_{\delta,\ell}$ in equation~\eqref{eq:piecewise-constant-hazard-pop-level}, we get that the maximum likelihood estimate of the event-specific hazard function is
\[
\widehat{\lambda}_\delta(t)
=\begin{cases}
{\displaystyle \frac{d_{\delta,\ell}}{(t_\ell\!-\!t_{\ell-1})n_\ell}} & \!\text{if }t_\ell\in(t_{\ell-1},t_\ell]\text{ for }\ell\in[L],\\
0 & \!\text{otherwise}.
\end{cases}
\]
We can then integrate to get the \emph{event-specific cumulative hazard function}:
\begin{align*}
\widehat{\Lambda}_\delta(t_\ell)
&\triangleq
   \int_0^{t_\ell}\widehat{\lambda}_\delta(u)\textrm{d}u\\
&=\sum_{a=1}^\ell\frac{d_{\delta,a}}{(t_a-t_{a-1})n_a}(t_a-t_{a-1})\\
&=\sum_{a=1}^\ell\frac{d_{\delta,a}}{n_a}.
\end{align*}
Note that the equation above is evaluating $\widehat{\Lambda}_\delta(t)$ specifically at time points along the discrete time grid $t_\ell$ for $\ell\in[L]$. When we evaluate at times in between these discrete grid points, the integration does constant-hazard interpolation. In particular, for any time $t\in(t_{\ell-1},t_\ell)$ (with $\ell\in[L]$), we have
\[
\widehat{\Lambda}_\delta(t)=\sum_{a=1}^{\ell-1}\frac{d_{\delta,a}}{n_a}+\frac{d_{\delta,\ell}}{(t_\ell-t_{\ell-1})n_\ell}(t-t_{\ell-1}),
\]
with the convention that when $\ell=1$, the first term on the right-hand side is 0.

However, at this point, to obtain the actual commonly used version of the Aalen-Johansen estimator stated in equation~\eqref{eq:simplified-AJ}, instead of doing this constant-hazard interpolation, we do a ``forward-filling'' interpolation. Namely, we approximate the event-specific cumulative hazard function with
\begin{align*}
&\widehat{\Lambda}_\delta^{\text{forward-fill}}(t) \\
&\quad\triangleq
  \begin{cases}
    0
    & \text{if }t=0 \\
    {\displaystyle \sum_{a=1}^\ell\frac{d_{\delta,a}}{n_a}}
    & \text{if }t\in(t_{\ell-1},t_\ell]\text{ for }\ell\in[L], \\
    {\displaystyle \sum_{a=1}^L\frac{d_{\delta,a}}{n_a}}
    & \text{if }t>t_L.
  \end{cases}
\end{align*}
A similar derivation can be used to show that the classical Kaplan-Meier estimator is a forward-filled interpolated version of a maximum likelihood estimator (for a derivation of this result, see Example~2.6 and Section~2.A.3 of \citet{chen2024introduction}; note that we basically reduce the setup to a standard survival analysis setup where there is a single critical event, corresponding to any of the $m$ critical events happening).

We now finally use the main idea of the Aalen-Johansen estimator \citep{aalen1978empirical}. For reference, the general Aalen-Johansen estimator for multistate survival analysis is given by the matrix product in equation~(3.17) of \citet{cook2018multistate}. We instead use the special case given by equation~(4.3) in Cook and Lawless's book, which states that, using Riemann-Stieltjes integral notation, we can estimate $F_\delta^{\text{pop}}(t)={\mathbb{P}(T\le t,\Delta^*=\delta)}$ with
\[
\int_0^t \widehat{S}(u^-) \textrm{d}\widehat{\Lambda}_\delta(u),
\]
where $\widehat{S}$ is any estimator of the survival function $S(t)=\mathbb{P}(T>t)$, and $\widehat{\Lambda}_\delta$ is any estimator of the event-specific cumulative hazard function $\Lambda_\delta$ (these functions are all at the population level and are not conditioned on a feature vector); ``$\widehat{S}(u^-)$'' refers to evaluating $\widehat{S}$ at the time immediately before $u$. When we plug in $\widehat{S}^\text{KM}$ (from equation~\eqref{eq:KM}) for $\widehat{S}$ and $\widehat{\Lambda}_\delta^{\text{forward-fill}}$ for~$\widehat{\Lambda}_\delta$, this Riemann-Stieltjes integral can be evaluated in closed-form and is precisely equal to
\[
\widehat{F}_\delta^{\text{AJ}}(t)
\triangleq \sum_{\ell=1}^L
  \widehat{S}^{\text{KM}}(t_{\ell-1})
  \frac{\ind\{t_\ell \le t\}d_{\delta,\ell}}
       {n_\ell}.
\]
This completes the proof and also indicates the interpolation used (forward-filling interpolation).$\hfill\blacksquare$

\section{DKAJ Log Likelihood Calculation}
\label{apd:DKAJ-log-likelihood}

The derivation is similar to equation~\eqref{eq:AJ-log-likelihoood-calculation}. Using equations~\eqref{eq:likelihood}, \eqref{eq:piecewise-constant-hazard}, and \eqref{eq:psi-detailed}, we have
\begin{align*}
&\log\mathcal{L} \\
&\!=
  \sum_{\delta=1}^m
    \sum_{i=1}^n\!
      \Bigg[
        \ind\{\Delta_i\!=\!\delta\}\log \lambda_\delta(Y_i|X_i)
        \!-\!
        \int_0^{Y_i}\!\!\lambda_\delta(u|X_i)\textrm{d}u
      \Bigg] \\
&=\sum_{\delta=1}^m
    \sum_{i=1}^n\!
      \bigg[
        \ind\{\Delta_i\!=\!\delta\}\log\frac{\psi_{\delta,\kappa(Y_i)}(X_i;\theta)}{t_{\kappa(Y_i)}-t_{\kappa(Y_i)-1}}\\
&\phantom{
   =\sum_{\delta=1}^m
     \sum_{i=1}^n\!
       \Bigg[~
 }
        -
        \sum_{\ell=1}^{\kappa(Y_i)}
          \frac{\psi_{\delta,\ell}(X_i;\theta)}{t_\ell-t_{\ell-1}}(t_\ell-t_{\ell-1})
      \bigg] \\
&=\sum_{\delta=1}^m
    \sum_{i=1}^n\!
      \bigg[
        \ind\{\Delta_i\!=\!\delta\}\log\frac{\psi_{\delta,\kappa(Y_i)}(X_i;\theta)}{t_{\kappa(Y_i)}-t_{\kappa(Y_i)-1}}\\
&\phantom{
   =\sum_{\delta=1}^m
     \sum_{i=1}^n\!
       \Bigg[~
 }
        -
        \sum_{\ell=1}^{\kappa(Y_i)}
          \psi_{\delta,\ell}(X_i;\theta)
      \bigg] \\
&=\sum_{\delta=1}^m
    \sum_{i=1}^n\!
      \bigg[
        \ind\{\Delta_i\!=\!\delta\}
        \log\psi_{\delta,\kappa(Y_i)}(X_i;\theta)\\
&\phantom{
   =\sum_{\delta=1}^m
     \sum_{i=1}^n\!
       \Bigg[~
 }
        -
        \sum_{\ell=1}^{\kappa(Y_i)}
          \psi_{\delta,\ell}(X_i;\theta)
      \bigg] \\
&\phantom{
   =~
 }
  -
  \underbrace{
    \sum_{\delta=1}^m
      \sum_{i=1}^n\!
        \ind\{\Delta_i\!=\!\delta\}\log(t_{\kappa(Y_i)}-t_{\kappa(Y_i)-1})
  }_{\text{constant w.r.t.~neural net parameters~}\theta}.
\end{align*}
Lastly, to emphasize the dependence of the log likelihood on $\theta$, we use the notation ``$\log\mathcal{L}(\theta)$'' instead of just ``$\log\mathcal{L}$''.

\section{DeepHit Loss}
\label{apd:deephit-loss}

DeepHit \citep{lee2018deephit} uses both a negative log likelihood loss and a ranking loss. The negative log likelihood loss is precisely $\mathbb{L}_{\text{NLL}}=-\frac{1}{n}\log\mathcal{L}^{\text{train}}$, where $\log\mathcal{L}^{\text{train}}$ is given in equation~\eqref{eq:DKAJ-log-likelihood-train}. For hyperparameter $\sigma>0$, the DeepHit ranking loss can be stated as
\[
\mathbb{L}_{\text{ranking}}
\triangleq
  \frac{1}{n^2}
  \sum_{\delta=1}^m
    \sum_{i=1}^n
      \sum_{j=1}^n
        \ind\{\Delta_i=\delta,Y_i<Y_j\}
          \Xi_{\delta,i,j}
\]
where
\[
\Xi_{\delta,i,j}
\triangleq
  \exp\Big(\frac{\widehat{F}_\delta^{\text{DKAJ}}(Y_i|X_j) - \widehat{F}_\delta^{\text{DKAJ}}(Y_i|X_i)}{\sigma}\Big).
\]
Note that the scale factor we use for the ranking loss is different from the original DeepHit paper but corresponds to the implementation in the now-standard \texttt{pycox} software package by \citet{kvamme2019time}. %
The overall DeepHit loss, following the \texttt{pycox} implementation, is
\[
\mathbb{L}_{\text{DeepHit}}
\triangleq
  \alpha \mathbb{L}_{\text{NLL}} + (1-\alpha)\mathbb{L}_{\text{ranking}},
\]
where $\alpha\in[0,1]$ is a hyperparameter.

\section{TUNA Warm-up with SurvivalBoost}
\label{apd:tuna-survivalboost}

For warm-starting neural net training for large datasets, the survival kernets paper \citep{chen2024survival} found that using a decision tree ensemble method that scales to large datasets is very helpful in practice (in a warm-start procedure called Tree ensemble Under a Neural Approximation (TUNA)). We wanted to adopt this TUNA warm-start strategy but the tree ensemble model that survival kernets used (XGBoost) does not, to the best of our knowledge, support competing risks. In fact, the only competing risks decision tree ensemble model we know of that scales to large datasets was only recently developed (SurvivalBoost by \cite{alberge2025survival}). Our method uses SurvivalBoost to warm-start neural net training and not XGBoost. We point out two implementation details needed to get the TUNA warm-start with SurvivalBoost to work:
\begin{itemize}[leftmargin=*]
    \item First, the warm-start procedure needs to know predicted leaf IDs per decision tree in the tree ensemble. At the time of writing, this functionality is not provided by SurvivalBoost, so we coded this part ourselves, building off the SurvivalBoost codebase.
    \item TUNA relies on approximating the kernel function of the base tree ensemble model (for two data points, the kernel function value is the fraction of trees that place the two data points in the same leaf). However, for SurvivalBoost, whether two data points are in the same leaf for a specific tree is time-dependent (i.e., at two different evaluation times $t_1$ and $t_2$, it is possible that two data points are considered to be in the same leaf at time $t_1$ but not at time $t_2$). Under the hood, SurvivalBoost treats time as an additional feature (i.e., time is appended to the feature vector per data point) when training a tree ensemble. This complication did not arise in the original survival kernets paper when it used XGBoost to warm-start (the XGBoost survival models don't have this time-dependent behavior as to which data points are considered to be in the same leaf). To deal with this, we randomly sample 10 evaluation times per training patient during the warm-start procedure.
\end{itemize}

\section{Summary Function Fine-Tuning}
\label{apd:SFT}

We now treat $d_{\delta,\ell}^{\text{cluster}}(q)$ and $n_\ell^{\text{cluster}}(q)$ as neural nets initialized to the values specified in equations~\eqref{eq:cluster-event-count} and~\eqref{eq:cluster-at-risk-count}. Specifically, we set
\begin{align*}
d_{\delta,\ell}^{\text{cluster}}(q)
&= \exp(\gamma_{q,\delta,\ell}) + \exp(\gamma_{\delta,\ell}^{\text{baseline}}),
\end{align*}
where $\gamma_{q,\delta,\ell}\in\mathbb{R}$ and $\gamma_{\delta,\ell}^{\text{baseline}}\in\mathbb{R}$ are unconstrained neural net parameters; we collect these particular neural net parameters all into the variable $\gamma$. This means that $\gamma$ consists of:
\begin{itemize}[leftmargin=*,itemsep=0pt]

\item $\gamma_{q,\delta,\ell}$ for $q\in\mathcal{Q}$, $\delta\in[m]$, and $\ell\in[L]$,

\item $\gamma_{\delta,\ell}^{\text{baseline}}$ for $\delta\in[m]$, and $\ell\in[L]$.

\end{itemize}
We initialize each $\gamma_{\delta,\ell}^{\text{baseline}}$ to be a big negative value (e.g., $\gamma_{\delta,\ell}^{\text{baseline}}=\log(10^{-12})=-27.631\dots$), so that $\exp(\gamma_{\ell}^{\text{baseline}})=10^{-12}$. Meanwhile, we initialize each neural net parameter
\[
\gamma_{q,\delta,\ell}
= \log\Big(\sum_{j\in\mathcal{C}_q} \ind\{\Delta_j=\delta, Y_j=t_\ell\}\Big).
\]
Thus, at these initial values,
$d_{\delta,\ell}^{\text{cluster}}(q)$ is approximately equal to $\sum_{j\in\mathcal{C}_q} \ind\{\Delta_j=\delta, Y_j=t_\ell\}$ (corresponding to equation~\eqref{eq:cluster-event-count}).

Next, we explain how we model $n_\ell^{\text{cluster}}(q)$. Importantly, we do not parameterize this function directly. Instead, we define a new summary function corresponding to a pseudocount for how many points are censored at time~$\ell$ for cluster~$q$. Specifically, we define
\begin{align*}
c_{\ell}^{\text{cluster}}(q)
&= \exp(\omega_{q,\ell}) + \exp(\omega_{\ell}^{\text{baseline}}),
\end{align*}
where $\omega_{q,\ell}\in\mathbb{R}$ and $\omega_{\ell}^{\text{baseline}}\in\mathbb{R}$ are unconstrained neural net parameters; we collect these parameters into the variable $\omega$. This means that $\omega$ consists of:
\begin{itemize}[leftmargin=*,itemsep=0pt]

\item $\omega_{q,\ell}$ for $q\in\mathcal{Q}$ and $\ell\in[L]$,

\item $\omega_\ell^{\text{baseline}}$ for $\ell\in[L]$.

\end{itemize}
We have the recurrence relation
\[
n_\ell^{\text{cluster}}(q) = \sum_{\delta=1}^m d_{\delta,\ell}^{\text{cluster}}(q) + c_{\ell}^{\text{cluster}}(q) + n_{\ell+1}^{\text{cluster}}(q),
\]
where $n_{L+1}^{\text{cluster}}(q)\triangleq0$. Thus, we would like to initialize neural net parameters $\omega_{q,\ell}$ and $\omega_{\ell}^{\text{baseline}}$ as to (approximately) satisfy
\begin{align*}
c_{\ell}^{\text{cluster}}(q)
&= n_\ell^{\text{cluster}}(q) - n_{\ell+1}^{\text{cluster}}(q) - \sum_{\delta=1}^m d_{\delta,\ell}^{\text{cluster}}(q) \\
&= \sum_{j\in\mathcal{C}_q} \ind\{Y_j\ge t_\ell\}
   -
   \sum_{j\in\mathcal{C}_q} \ind\{Y_j\ge t_{\ell+1}\} \\
&\quad
   -
   \sum_{\delta=1}^m
     \sum_{j\in\mathcal{C}_q} \ind\{\Delta_j=\delta, Y_j=t_\ell\},
\end{align*}
where the second equality is from plugging in equations \eqref{eq:cluster-event-count} and~\eqref{eq:cluster-at-risk-count}.
It suffices to set $\omega_{\ell}^{\text{baseline}}$ to again be a large negative value, e.g., $\omega_{\ell}^{\text{baseline}} = \log(10^{-12}) = -27.631\dots$, so that $\exp(\omega_{\ell}^{\text{baseline}}) = 10^{-12}$. Meanwhile, we set
\begin{align*}
\omega_{q,\ell}
= \log\Bigg(
    &\sum_{j\in\mathcal{C}_q} \ind\{Y_j\ge t_\ell\} \\
    &-
     \sum_{j\in\mathcal{C}_q} \ind\{Y_j\ge t_{\ell+1}\} \\
    &-
     \sum_{\delta=1}^m
       \sum_{j\in\mathcal{C}_q} \ind\{\Delta_j=\delta, Y_j=t_\ell\}
  \Bigg).
\end{align*}
To summarize, we have defined $d_{\delta,\ell}^{\text{cluster}}(q)$ and $c_{\ell}^{\text{cluster}}(q)$ as very simple neural nets with parameters $\gamma$ and $\omega$, respectively. From these we can back out the cluster-specific at risk count function
\[
n_\ell^{\text{cluster}}(q) = \sum_{\delta=1}^m d_{\delta,\ell}^{\text{cluster}}(q) + c_{\ell}^{\text{cluster}}(q) + n_{\ell+1}^{\text{cluster}}(q).
\]
Note that $n_\ell^{\text{cluster}}(q)$ depends on both $\gamma$ and $\omega$.

At this point, we treat the kernel function $K$ as fixed, meaning that the neural net it depends on ($f$ with parameters stored in $\widehat{\theta}$) is fixed. We also treat the cluster assignment as fixed. Then for any data point with feature vector $x$, its predicted event counts and at-risk counts are given by equations~\eqref{eq:DKAJ-event-count} and~\eqref{eq:DKAJ-at-risk-count}, which now are in terms of the neural net functions we just described with parameters $\gamma$ and $\omega$; we reproduce these below for the reader's convenience, where we now emphasize the dependence on $\gamma$ and $\omega$:
\begin{align*}
d_{\delta,\ell}^{\text{DKAJ}}(x;\gamma)
&\triangleq
  \sum_{q\in\mathcal{Q}(x;\tau)} K(x,X_q)d_{\delta,\ell}^{\text{cluster}}(q;\gamma) \tag*{} \\
n_\ell^{\text{DKAJ}}(q;\gamma,\omega)
&\triangleq
  \sum_{q\in\mathcal{Q}(x;\tau)} K(x,X_q)n_\ell^{\text{cluster}}(q;\gamma,\omega)
\end{align*}
Combining these two equations with equations~\eqref{eq:psi} and~\eqref{eq:piecewise-constant-hazard}, followed by plugging the resulting event-specific hazard function into the log of the likelihood function in equation~\eqref{eq:likelihood}, we get
\begin{align}
\log\mathcal{L}
&=\sum_{\delta=1}^m
    \sum_{i=1}^n\!
      \bigg[
        \ind\{\Delta_i\!=\!\delta\}
        \log\frac{d_{\delta,\kappa(Y_i)}^{\text{DKAJ}}(X_i;\gamma)}{n_\ell^{\text{DKAJ}}(X_i;\gamma,\omega)} \nonumber\\
&\phantom{
   =\sum_{\delta=1}^m
     \sum_{i=1}^n\!
       \Bigg[~
 }
        -
        \sum_{\ell=1}^{\kappa(Y_i)}
          \frac{d_{\delta,\ell}^{\text{DKAJ}}(X_i;\gamma)}{n_\ell^{\text{DKAJ}}(X_i;\gamma,\omega)}
      \bigg]+\text{constant},
\label{eq:SFT-log-likelihood}
\end{align}
which we can maximize using standard minibatch gradient descent (in which we would instead minimize the negative log likelihood loss $\mathbb{L}_{\text{NLL}}\triangleq-\frac{1}{n}\log\mathcal{L}$). Note that we do not need the leave-one-out strategy described in Section~\ref{sec:neural-net-formulation} because the functions $d_{\delta,\ell}^{\text{DKAJ}}(x;\gamma)$ and $n_\ell^{\text{DKAJ}}(x;\gamma,\omega)$ are not parameterized in terms of summations over the $n$ training points (which had previously been the case for equations~\eqref{eq:kernel-event-count} and~\eqref{eq:kernel-at-risk-count}). Adding a ranking loss (as to obtain the full DeepHit loss) is also possible; the idea is the same as what is presented in \appendixref{apd:deephit-loss}.

\section{Experimental Details} \label{apd:exp}

\subsection{Datasets} 

In this section, we provide details of the datasets used in our experiments, including cohort selection criteria, feature definitions, and preprocessing. For all datasets, the preprocessing pipeline (encoding, scaling, imputation) is fit only on the training set and then applied consistently to the test set, to avoid data leakage. When applicable, for missing values on categorical and binary features, we use the mode of the corresponding features in the training for imputation; whereas average feature values in the training set is used for continuous variable imputation. Continuous variables are standardized, while categorical features are one-hot encoded.

\paragraph{PBC}
The PBC dataset is derived from a clinical trial of 1,945 patients with primary biliary cirrhosis (PBC). The full feature list we use in our modeling includes: 
\begin{itemize}[noitemsep,topsep=2pt]
  \item \texttt{D-penicil}: indicator for treatment with D-penicillamine.  
  \item \texttt{female}: binary sex indicator.  
  \item \texttt{ascites}: presence of ascites.  
  \item \texttt{hepatomegaly}: presence of hepatomegaly.  
  \item \texttt{spiders}: presence of spider angiomas.  
  \item \texttt{edema}: categorical indicator of edema (none, present without diuretics, present despite diuretics).  
  \item \texttt{histologic}: histologic stage of disease.  
  \item \texttt{serBilir}: serum bilirubin.  
  \item \texttt{serChol}: serum cholesterol.  
  \item \texttt{albumin}: serum albumin.  
  \item \texttt{alkaline}: alkaline phosphatase.  
  \item \texttt{SGOT}: serum glutamic-oxaloacetic transaminase.  
  \item \texttt{platelets}: platelet count.  
  \item \texttt{prothrombin}: prothrombin time.  
  \item \texttt{age}: patient age at study entry.  
\end{itemize}  

\paragraph{Framingham}
The Framingham dataset originates from the Framingham Heart Study and includes 4,434 participants. Although longitudinal data are available, we restrict to the earliest observed features for each patient. We excluded patients with missing event time or censoring indicator. The full feature list we use in our modeling includes:
\begin{itemize}[noitemsep,topsep=2pt]
  \item \texttt{SEX}: Binary indicator of sex (0 = male, 1 = female).  
  \item \texttt{CURSMOKE}: Current smoking status (1 = current smoker, 0 = non-smoker).  
  \item \texttt{DIABETES}: Indicator of physician-diagnosed diabetes mellitus.  
  \item \texttt{BPMEDS}: Indicator for use of antihypertensive medications at baseline.  
  \item \texttt{PREVCHD}: History of coronary heart disease (CHD) prior to baseline.  
  \item \texttt{PREVAP}: History of angina pectoris prior to baseline.  
  \item \texttt{PREVMI}: History of myocardial infarction prior to baseline.  
  \item \texttt{PREVSTRK}: History of stroke prior to baseline.  
  \item \texttt{PREVHYP}: History of physician-diagnosed hypertension.  
  \item \texttt{educ}: Educational attainment, coded as a categorical variable with four levels from 1 to 4.  
  \item \texttt{TOTCHOL}: Total serum cholesterol (mg/dL).  
  \item \texttt{AGE}: Age in years at baseline.  
  \item \texttt{SYSBP}: Systolic blood pressure (mmHg).  
  \item \texttt{DIABP}: Diastolic blood pressure (mmHg).  
  \item \texttt{CIGPDAY}: Number of cigarettes smoked per day (self-reported).  
  \item \texttt{BMI}: Body mass index (kg/m\textsuperscript{2}).  
  \item \texttt{HEARTRTE}: Resting heart rate (beats per minute).  
  \item \texttt{GLUCOSE}: Fasting blood glucose level (mg/dL).  
\end{itemize}

\paragraph{SEER}
The SEER dataset is obtained from the U.S. Surveillance, Epidemiology, and End Results Program. While SEER spans multiple decades and includes diverse cancer types and both sexes, in this work we restrict to female patients diagnosed with breast cancer in the year 2010. For patients with multiple records, only the first entry per patient is retained. Patients without survival time are excluded. The outcome is breast cancer–specific mortality, with cardiovascular disease (CVD) mortality treated as the competing risk.  
The full feature list we use in our modeling includes:
  \begin{itemize}[noitemsep,topsep=0pt]
    \item \texttt{Race and origin recode}: patient race and ethnicity.  
    \item \texttt{Laterality}: laterality of the primary breast tumor (e.g., left, right).  
    \item \texttt{Diagnostic Confirmation}: method of diagnosis (e.g., histology, cytology).  
    \item \texttt{Histology recode}: histologic tumor classification.  
    \item \texttt{Chemotherapy recode}: indicator for chemotherapy administration.  
    \item \texttt{Radiation recode}: indicator for radiation therapy.  
    \item \texttt{ER Status Recode Breast Cancer}: estrogen receptor status.  
    \item \texttt{PR Status Recode Breast Cancer}: progesterone receptor status.  
    \item \texttt{Sequence number}: sequence of the tumor relative to prior malignancies.  
    \item \texttt{RX Summ--Surg Prim Site}: type of surgery performed on the primary tumor.  
    \item \texttt{CS extension}: clinical extent of primary tumor spread.  
    \item \texttt{CS lymph nodes}: involvement of regional lymph nodes.  
    \item \texttt{CS mets at dx}: presence of distant metastasis at diagnosis.  
    \item \texttt{Origin recode NHIA}: Hispanic origin classification.  
    \item \texttt{Grade Recode}: histologic grade (I–IV or unknown).  
    \item \texttt{Age recode with <1 year olds}: patient age grouped into categories (e.g., 01–04, 05–09, …, 85+).  
    \item \texttt{Year of diagnosis}: restricted to 2010 in this study.  
    \item \texttt{Total number of in situ/malignant tumors for patient}.  
    \item \texttt{Total number of benign/borderline tumors for patient}.  
    \item \texttt{CS tumor size}: tumor size in millimeters.  
    \item \texttt{Regional nodes examined}: number of lymph nodes examined.  
    \item \texttt{Regional nodes positive}: number of positive lymph nodes.  
  \end{itemize}

\paragraph{Synthetic}
We use the publicly available synthetic competing risks dataset by \citet{lee2018deephit}.\footnote{\url{https://github.com/chl8856/DeepHit}}

\subsection{Evaluation Metrics} \label{apd:eval-metric}
As we will soon see, both the integrated Brier score and time-dependent concordance index rely on a time grid.
To ensure consistent comparison across models, evaluation is performed on a common time grid within a dataset across different data splits and hyperparameter configurations. For the evaluation time grid, we use 100 evenly spaced quantiles of the observed event times (see \citet{kvamme2021continuous}, Section~3.1), starting from the minimum observed event times and truncated at the 90th percentile to reduce instability at the tail. This grid is constructed from the entire dataset for each experiment, instead of just from the training set, to make sure model comparison across different data splits is consistent. For discrete-time competing risks models (e.g., DeepHit and DKAJ), linear interpolation of the predicted CIFs is applied in order to evaluate at arbitrary time points.

\paragraph{Competing risks Brier score (BS)} (\citet{gerds2021medical}, Section 5.4.2)
Let $\delta \in [m]$ index the competing risks and $t \geq 0$. Given a cumulative incidence function (CIF) estimator $\widehat{F}_\delta(\cdot|x)$, the Brier score for event $\delta$ at time $t$ is defined as
\begin{align*}
\text{BS}_\delta(t)
:=
&  \frac{1}{n}
    \sum_{i=1}^n
      \bigg[
       \frac{(1 - \widehat{F}_\delta(t|X_i))^2 \ind\{\Delta_i = \delta\} \ind\{ Y_i \le t\}}
             {\widehat{S}_{\text{censor}}(Y_i)} \\
&       +
        \frac{(\widehat{F}_\delta(t|X_i))^2 \ind\{\Delta_i \ne \delta,\, \Delta_i \ne 0\} \ind\{ Y_i \le t\}}
             {\widehat{S}_{\text{censor}}(Y_i)} \\
&       +
        \frac{(\widehat{F}_\delta(t|X_i))^2 \ind\{ Y_i > t \}}
             {\widehat{S}_{\text{censor}}(t)}
      \bigg],
\end{align*}
where $\widehat{S}_{\text{censor}}(\cdot)$ is the Kaplan-Meier estimate of the censoring distribution. Lower values indicate better performance. Notice that the competing risks Brier score is slightly different from the Brier score under the classic single-event survival analysis.

\paragraph{Competing risks integrated Brier score (IBS)}  
To summarize performance across time, the competing risks Brier score is integrated over the evaluation time grid, producing the integrated Brier score (IBS). Lower values of IBS indicate better calibration and discrimination.

\paragraph{Time-dependent concordance index ($C^{\text{td}}$)}  
We also report the time-dependent concordance index \citep{antolini2005time}, which measures a model’s discriminative ability by assessing whether individuals with shorter survival times are assigned higher risk scores. When we are evaluating a model's performance for some event of interest using the $C^{\text{td}}$, we treat the occurrence of other events as censoring (i.e., the event indicator becomes censoring, and the event time of other events becomes the censoring time), which is a common strategy adopted in the literature. Higher values of $C^{\text{td}}$ indicate better discriminative performance.

\medskip  
In summary, IBS (lower is better) emphasizes both calibration and discrimination, whereas $C^{\text{td}}$ (higher is better) focuses only on discrimination. Both metrics are evaluated on the same truncated time grid described above.

\subsection{Hyperparameter Grids and Optimization Details} \label{apd:exp-hyperparam}

We perform hyperparameter selection via grid search, with model selection based on the best validation performance (either average $C^{\text{td}}$ or IBS across all events; see \appendixref{apd:eval-metric}). 

\paragraph{Optimization and early stopping}
For deep models (DeepHit, DSM, NeuralFG, DKAJ) we use a maximum of 1000 epochs with early stopping (patience 10).
DSM and NeuralFG (code from \texttt{NeuralFineGray}\footnote{\url{https://github.com/Jeanselme/NeuralFineGray/}}) stop on the validation objective.
DeepHit and DKAJ stop on the validation metric (IBS or \Ctd) aligned with the selection criterion. Appendix~\ref{apd:alt-es} reports results when DeepHit and DKAJ also stop on validation objective; conclusions are unchanged.
SurvivalBoost does not support early stopping in the reference implementation at the time of running experiments in this paper.
Batch size is 1024 for all neural models (not applicable to Fine-Gray, cs-Cox, RSF-CR, and SurvivalBoost).

For each method, we provide additional details for replicability as follows:

\paragraph{Fine and Gray}
We use the implementation provided in the \texttt{cmprsk}\footnote{\url{https://cran.r-project.org/web/packages/cmprsk}} package written in R, which is called in Python via the \texttt{rpy2} package. There are no tunable hyperparameters.

\paragraph{Cause-specific Cox PH (cs-Cox)}
We use the \texttt{CSC} module implementation provided in the \texttt{riskRegression}\footnote{\url{https://cran.r-project.org/web/packages/riskRegression}} package written in R, which is called in Python via the \texttt{rpy2} package. Specifically, glmnet with 5-fold cross-validation on partial log-likelihood loss is applied to select the penalty.

\paragraph{Random survival forest for competing risks (RSF-CR)}
We use the \texttt{rfsrc} module implemented in the \texttt{randomForestSRC}\footnote{\url{www.randomforestsrc.org}} package written in R, which is called in Python via the \texttt{rpy2} package.
\begin{itemize}[itemsep=-4pt,topsep=1pt]
    \item Number of trees: $\{500, 750, 1000\}$
    \item Number of variables to possibly split at each node: $\{\text{None}, 8, 16\}$
    \item Minimum size of terminal node: $\{15, 30\}$
    \item Number of grid of time points over the observed event times: $\{0, 64, 128\}$, where $0$ means using all observed event times
\end{itemize}  
For the hyperparameters not mentioned above, default values provided by the package are applied.

\paragraph{DeepHit}    
\begin{itemize}[itemsep=-4pt,topsep=1pt]
  \item Number of layers: $\{2, 4\}$  
  \item Hidden units per layer: $\{64, 128\}$  
  \item Learning rate: $\{0.01, 0.001\}$  
  \item $\alpha$: $\{0, 0.001, 0.01\}$  
  \item $\sigma$: $\{0.1, 1\}$  
  \item Number of time steps $L$: $\{0, 64, 128\}$
\end{itemize}
The number of time steps being 0 means we use all possible observed event times in the training set, but an upper limit of 512 is used to prevent the neural net architecture from being too complex.
When discretizing time into $k$ steps, we use evenly spaced quantiles of the observed event times (see \citet{kvamme2021continuous}, Section 3.1). This can lead to fewer unique time bins if many event times coincide.

\paragraph{Deep Survival Machines (DSM)}
\begin{itemize}[itemsep=-4pt,topsep=1pt]
  \item Number of layers: $\{2, 4\}$  
  \item Hidden units per layer: $\{64, 128\}$  
  \item Learning rate: $\{0.01, 0.001\}$  
  \item Number of mixture components $k$: $\{4, 8\}$  
  \item Survival distribution family: \{Weibull, LogNormal\}
  \item Discount factor: $\{0.5, 1\}$  
\end{itemize}  

\paragraph{Neural Fine-Gray (NeuralFG)}  
\begin{itemize}[itemsep=-4pt,topsep=1pt]
  \item Number of encoder layers: $\{2, 4\}$  
  \item Hidden units per encoder layer: $\{64, 128\}$  
  \item Number of survival-specific layers: $\{1\}$  
  \item Hidden units per survival-specific layer: $\{64, 128\}$  
  \item Learning rate: $\{0.01, 0.001\}$  
  \item Dropout: $\{0, 0.25, 0.5, 0.75\}$  
  \item Activation: $\tanh$  
\end{itemize}  

\paragraph{SurvivalBoost}  
\begin{itemize}[itemsep=-4pt,topsep=1pt]
  \item Learning rate: $\{0.01, 0.05, 0.1, 0.5\}$  
  \item Number of boosting iterations: $\{20, 100, 200\}$  
  \item Maximum tree depth: $\{-1, 4, 8, 16\}$  
  \item Number of time grid steps: $\{64, 128\}$  
  \item IPCW strategy: $\{\text{alternating}, \text{Kaplan--Meier}\}$  
\end{itemize}  

\paragraph{DKAJ}  
\begin{itemize}[itemsep=-4pt,topsep=1pt]
  \item Number of layers: $\{2, 4\}$  
  \item Hidden units per layer: $\{64, 128\}$  
  \item Learning rate: $\{0.01, 0.001\}$  
  \item $\alpha$: $\{0, 0.001, 0.01\}$  
  \item $\sigma$: $\{0.1, 1\}$  
  \item Number of time steps $L$: $\{0, 64, 128\}$  
  \item $\gamma$: $\{0\}$  
  \item $\beta$: $\{0.25, 0.5\}$  
  \item Squared radius: $\{0.1\}$  
  \item Minimum kernel weight (transformation of $\tau$): $\{10^{-2}\}$  
\end{itemize}
The number of time steps being 0 means we use all possible observed event times in the training set, but an upper limit of 512 is used to prevent the neural net architecture from being too complex. Moreover, we sample 10 random evaluation times per data point when using the TUNA neural net warm-start with SurvivalBoost (which leaf a data point is in per tree for SurvivalBoost depends on the evaluation time; we randomly choose different evaluation times). 

Following \citet{chen2024survival}, $\tau$ is parameterized as a transformation of the minimum kernel weight, the smallest contribution $K(x, q)$ can make for any exemplar $q$. 
Specifically, $\tau = \sqrt{-\log(\text{min kernel weight})}$.
In all experiments, we set the minimum kernel weight to $10^{-2}$, consistent with prior work in survival kernels \citep{chen2024survival}, which provides a reasonable trade-off between smoothness and locality across datasets.

We use the TUNA warm start strategy to initialize DKAJ. %
In doing this warm start, we (i) sample 10 random evaluation times per individual when pairing with SurvivalBoost (its leaf assignment depends on the evaluation time), and (ii) perform a small inner sweep over architectural and cluster parameters only for initialization, e.g., number of layers and hidden units, squared radius, and batch size. The best warm-start setting under the validation criterion is then frozen, and the main DKAJ likelihood training proceeds with the outer grid listed above. This two-stage scheme reduces the effective search space during the main optimization. %

\paragraph{DKAJ Summary Function Fine-tuning (SFT)}
\begin{itemize}[itemsep=-4pt,topsep=1pt]
    \item Fine-tuning learning rate: $\{0.01, 0.001, 0.0001\}$  
\end{itemize}
Note that we only use the model with SFT if it results in improved average validation performance across different events; otherwise, we backtrack to the original DKAJ model before SFT.

\section{Additional Experimental Results}

\subsection{Detailed Results Under Different Model Selection Criteria} \label{apd:exp-results}
Tables~\ref{tab:performance-tune-by-ibs} and \ref{tab:performance-tune-by-ctd} report the full test performance of all models under two different model selection criteria: the best validation IBS and the best validation $C^{\text{td}}$, both of which are averaged across events, respectively. As discussed in footnote~\ref{foot:validation-criterion} in Section~\ref{sec:experiments}, the choice of validation criterion has a substantial impact on the observed test performance. When models are tuned to minimize average IBS across events, the resulting $C^{\text{td}}$ scores are typically lower (Table~\ref{tab:performance-tune-by-ibs}); conversely, tuning to maximize average $C^{\text{td}}$ across events often yields inferior IBS values (Table~\ref{tab:performance-tune-by-ctd}). This trade-off is consistent across datasets and event types, illustrating the tension between an evaluation metric that accounts for calibration (IBS) vs one that focuses only on discrimination ($C^{\text{td}}$) that is well documented in survival analysis.

We also investigate the effect of applying the optional summary function fine-tuning (SFT; Section~\ref{sec:DKAJ-overall}) after training the DKAJ models. The corresponding results are reported in Tables~\ref{tab:performance-tune-by-ibs} and \ref{tab:performance-tune-by-ctd}. Overall, we observe that SFT often leads to further performance improvements for DKAJ across most datasets and event types. However, the magnitude of the improvement can be modest in some cases, and in rare instances, SFT slightly degrades performance on the test set.

\begin{table*}[htbp]
    \centering
    \floatconts {tab:performance-tune-by-ibs}
    {\caption{Test set IBS and $C^{\text{td}}$ (mean$~\!\pm~\!$std.~dev.~across 10 random splits) for primary and competing events (best score in \textbf{bold}, 2nd best in \blue{blue}). Model selection done by the best validation IBS; the IBS columns are shaded to highlight the results.}}
    {
    \begin{adjustbox}{max width=0.8\textwidth}
    \begin{tabular}{cccccc}
        \toprule
        \bfseries \multirow{2}{*}{Dataset} & \bfseries \multirow{2}{*}{Method} & \multicolumn{2}{c}{\cellcolor{gray!20}\bfseries IBS $(\downarrow)$}  & \multicolumn{2}{c}{\bfseries $C^{\text{td}}$ $(\uparrow)$} \\
        ~ & ~ & \cellcolor{gray!20}Primary & \cellcolor{gray!20}Competing & Primary & Competing \\
        \midrule
        \multirow{9}{*}{\rotatebox{90}{PBC}} & Fine-Gray & \cellcolor{gray!20}0.1086$\pm$0.0047 & \cellcolor{gray!20}0.0401$\pm$0.0046 & 0.8212$\pm$0.0119 & 0.8769$\pm$0.0203 \\
        ~ & cs-Cox & \cellcolor{gray!20}0.1052$\pm$0.0051 & \cellcolor{gray!20}0.0391$\pm$0.0047 & 0.8295$\pm$0.0102 & \blue{0.9094$\pm$0.0132} \\
        ~ & RSF-CR & \cellcolor{gray!20}\blue{0.0989$\pm$0.0035} & \cellcolor{gray!20}0.0412$\pm$0.0049 & \blue{0.8584$\pm$0.0082} & 0.8803$\pm$0.0143 \\
        ~ & DeepHit & \cellcolor{gray!20}0.1259$\pm$0.0121 & \cellcolor{gray!20}0.0552$\pm$0.0074 & 0.8243$\pm$0.0275 & 0.8050$\pm$0.0765 \\
        ~ & DSM & \cellcolor{gray!20}0.1103$\pm$0.0055 & \cellcolor{gray!20}0.0419$\pm$0.0041 & 0.8331$\pm$0.0103 & 0.8843$\pm$0.0204 \\
        ~ & NeuralFG & \cellcolor{gray!20}0.1055$\pm$0.0061 & \cellcolor{gray!20}0.0395$\pm$0.0056 & 0.8350$\pm$0.0151 & 0.9060$\pm$0.0133 \\
        ~ & SurvivalBoost & \cellcolor{gray!20}\bftab{0.0898$\pm$0.0046} & \cellcolor{gray!20}\bftab{0.0322$\pm$0.0041} & \bftab{0.8745$\pm$0.0071} & \bftab{0.9419$\pm$0.0072} \\
        ~ & DKAJ & \cellcolor{gray!20}0.1031$\pm$0.0071 & \cellcolor{gray!20}0.0394$\pm$0.0057 & 0.8242$\pm$0.0215 & 0.7885$\pm$0.0825 \\
        ~ & DKAJ + SFT & \cellcolor{gray!20}0.1037$\pm$0.0072 & \cellcolor{gray!20}\blue{0.0384$\pm$0.0058} & 0.8351$\pm$0.0177 & 0.8792$\pm$0.0444 \\
        \midrule
        \multirow{9}{*}{\rotatebox{90}{Framingham}} & Fine-Gray & \cellcolor{gray!20}0.0819$\pm$0.0037 & \cellcolor{gray!20}\bftab{0.0566$\pm$0.0035} & \blue{0.7733$\pm$0.0106} & \blue{0.7144$\pm$0.0182} \\
        ~ & cs-Cox & \cellcolor{gray!20}\bftab{0.0814$\pm$0.0039} & \cellcolor{gray!20}\bftab{0.0566$\pm$0.0034} & \bftab{0.7753$\pm$0.0105} & \bftab{0.7160$\pm$0.0173} \\
        ~ & RSF-CR & \cellcolor{gray!20}\blue{0.0815$\pm$0.0031} & \cellcolor{gray!20}0.0573$\pm$0.0036 & 0.7691$\pm$0.0097 & 0.6924$\pm$0.0122 \\
        ~ & DeepHit & \cellcolor{gray!20}0.0941$\pm$0.0049 & \cellcolor{gray!20}0.0645$\pm$0.0055 & 0.7099$\pm$0.0146 & 0.6464$\pm$0.0224 \\
        ~ & DSM & \cellcolor{gray!20}0.0847$\pm$0.0045 & \cellcolor{gray!20}0.0610$\pm$0.0046 & 0.7657$\pm$0.0154 & 0.7119$\pm$0.0183 \\
        ~ & NeuralFG & \cellcolor{gray!20}0.0880$\pm$0.0035 & \cellcolor{gray!20}0.0624$\pm$0.0039 & 0.7060$\pm$0.0706 & 0.5164$\pm$0.1516 \\
        ~ & SurvivalBoost & \cellcolor{gray!20}0.0824$\pm$0.0038 & \cellcolor{gray!20}0.0573$\pm$0.0036 & 0.7640$\pm$0.0136 & 0.6975$\pm$0.0158 \\
        ~ & DKAJ & \cellcolor{gray!20}0.0844$\pm$0.0036 & \cellcolor{gray!20}0.0578$\pm$0.0034 & 0.7613$\pm$0.0155 & 0.6870$\pm$0.0192 \\
        ~ & DKAJ + SFT & \cellcolor{gray!20}0.0853$\pm$0.0031 & \cellcolor{gray!20}0.0579$\pm$0.0035 & 0.7629$\pm$0.0158 & 0.6882$\pm$0.0187 \\
        \midrule
        \multirow{9}{*}{\rotatebox{90}{SEER}} & Fine-Gray & \cellcolor{gray!20}0.0637$\pm$0.0013 & \cellcolor{gray!20}\blue{0.0165$\pm$0.0010} & 0.8151$\pm$0.0021 & \blue{0.8478$\pm$0.0099} \\
        ~ & cs-Cox & \cellcolor{gray!20}0.0638$\pm$0.0013 & \cellcolor{gray!20}\bftab{0.0164$\pm$0.0009} & 0.7630$\pm$0.0085 & 0.8415$\pm$0.0132 \\
        ~ & RSF-CR & \cellcolor{gray!20}\blue{0.0599$\pm$0.0013} & \cellcolor{gray!20}0.0167$\pm$0.0010 & \blue{0.8346$\pm$0.0033} & 0.8244$\pm$0.0139 \\
        ~ & DeepHit & \cellcolor{gray!20}0.0721$\pm$0.0012 & \cellcolor{gray!20}0.0228$\pm$0.0017 & 0.8043$\pm$0.0140 & 0.8134$\pm$0.0494 \\
        ~ & DSM & \cellcolor{gray!20}0.0613$\pm$0.0011 & \cellcolor{gray!20}0.0171$\pm$0.0008 & 0.7809$\pm$0.0096 & 0.8407$\pm$0.0089 \\
        ~ & NeuralFG & \cellcolor{gray!20}0.0608$\pm$0.0014 & \cellcolor{gray!20}0.0166$\pm$0.0009 & 0.7736$\pm$0.0094 & 0.8360$\pm$0.0154 \\
        ~ & SurvivalBoost & \cellcolor{gray!20}\bftab{0.0584$\pm$0.0014} & \cellcolor{gray!20}\blue{0.0165$\pm$0.0008} & \bftab{0.8424$\pm$0.0029} & \bftab{0.8558$\pm$0.0100} \\
        ~ & DKAJ & \cellcolor{gray!20}0.0609$\pm$0.0011 & \cellcolor{gray!20}0.0167$\pm$0.0011 & 0.8277$\pm$0.0043 & 0.8250$\pm$0.0153 \\
        ~ & DKAJ + SFT & \cellcolor{gray!20}0.0792$\pm$0.0138 & \cellcolor{gray!20}0.0197$\pm$0.0041 & 0.8274$\pm$0.0038 & 0.8316$\pm$0.0182 \\
        \midrule
        \multirow{9}{*}{\rotatebox{90}{Synthetic}} & Fine-Gray & \cellcolor{gray!20}0.1851$\pm$0.0035 & \cellcolor{gray!20}0.1827$\pm$0.0027 & 0.5823$\pm$0.0051 & 0.5917$\pm$0.0071 \\
        ~ & cs-Cox & \cellcolor{gray!20}0.1851$\pm$0.0035 & \cellcolor{gray!20}0.1827$\pm$0.0027 & 0.5809$\pm$0.0051 & 0.5902$\pm$0.0073 \\
        ~ & RSF-CR & \cellcolor{gray!20}0.1745$\pm$0.0033 & \cellcolor{gray!20}0.1739$\pm$0.0027 & 0.7217$\pm$0.0076 & 0.7169$\pm$0.0040 \\
        ~ & DeepHit & \cellcolor{gray!20}0.1732$\pm$0.0046 & \cellcolor{gray!20}0.1702$\pm$0.0026 & 0.7214$\pm$0.0181 & 0.7233$\pm$0.0127 \\
        ~ & DSM & \cellcolor{gray!20}0.1743$\pm$0.0034 & \cellcolor{gray!20}0.1715$\pm$0.0028 & 0.7255$\pm$0.0067 & 0.7295$\pm$0.0044 \\
        ~ & NeuralFG & \cellcolor{gray!20}\bftab{0.1631$\pm$0.0033} & \cellcolor{gray!20}\bftab{0.1606$\pm$0.0027} & \bftab{0.7479$\pm$0.0069} & \bftab{0.7518$\pm$0.0041} \\
        ~ & SurvivalBoost & \cellcolor{gray!20}0.1734$\pm$0.0034 & \cellcolor{gray!20}0.1712$\pm$0.0025 & 0.7062$\pm$0.0083 & 0.7120$\pm$0.0034 \\
        ~ & DKAJ & \cellcolor{gray!20}\blue{0.1672$\pm$0.0039} & \cellcolor{gray!20}\blue{0.1647$\pm$0.0029} & \blue{0.7318$\pm$0.0083} & \blue{0.7346$\pm$0.0052} \\
        ~ & DKAJ + SFT & \cellcolor{gray!20}0.1777$\pm$0.0086 & \cellcolor{gray!20}0.1729$\pm$0.0056 & 0.7272$\pm$0.0086 & 0.7310$\pm$0.0060 \\
        \bottomrule
    \end{tabular}
    \end{adjustbox}
    }
\end{table*}

\begin{table*}[htbp]
    \centering
    \floatconts {tab:performance-tune-by-ctd}
    {\caption{Test set IBS and $C^{\text{td}}$ (mean$~\!\pm~\!$std.~dev.~across 10 random splits) for primary and competing events (best score in \textbf{bold}, 2nd best in \blue{blue}). Model selection done by the best validation $C^{\text{td}}$; the \Ctd~columns are shaded to highlight the results.}}
    {
    \begin{adjustbox}{max width=0.8\textwidth}
    \begin{tabular}{cccccc}
        \toprule
        \bfseries \multirow{2}{*}{Dataset} & \bfseries \multirow{2}{*}{Method} & \multicolumn{2}{c}{\bfseries IBS $(\downarrow)$}  & \multicolumn{2}{c}{\cellcolor{gray!20}\bfseries \Ctd~$(\uparrow)$} \\
        ~ & ~ & Primary & Competing & \cellcolor{gray!20}Primary & \cellcolor{gray!20}Competing \\
        \midrule
        \multirow{9}{*}{\rotatebox{90}{PBC}} & Fine-Gray & 0.1086$\pm$0.0047 & 0.0401$\pm$0.0046 & \cellcolor{gray!20}0.8212$\pm$0.0119 & \cellcolor{gray!20}0.8769$\pm$0.0203 \\
        ~ & cs-Cox & 0.1052$\pm$0.0051 & 0.0391$\pm$0.0047 & \cellcolor{gray!20}0.8295$\pm$0.0102 & \cellcolor{gray!20}0.9094$\pm$0.0132 \\
        ~ & RSF-CR & \blue{0.0995$\pm$0.0035} & 0.0414$\pm$0.0048 & \cellcolor{gray!20}\blue{0.8584$\pm$0.0089} & \cellcolor{gray!20}0.8746$\pm$0.0163 \\
        ~ & DeepHit & 0.1308$\pm$0.0052 & 0.0785$\pm$0.0086 & \cellcolor{gray!20}0.8407$\pm$0.0131 & \cellcolor{gray!20}0.9060$\pm$0.0130 \\
        ~ & DSM & 0.1185$\pm$0.0099 & 0.0404$\pm$0.0049 & \cellcolor{gray!20}0.8319$\pm$0.0115 & \cellcolor{gray!20}0.9039$\pm$0.0189 \\
        ~ & NeuralFG & 0.1060$\pm$0.0058 & \blue{0.0390$\pm$0.0046} & \cellcolor{gray!20}0.8363$\pm$0.0150 & \cellcolor{gray!20}\blue{0.9120$\pm$0.0149} \\
        ~ & SurvivalBoost & \bftab{0.0924$\pm$0.0085} & \bftab{0.0329$\pm$0.0050} & \cellcolor{gray!20}\bftab{0.8719$\pm$0.0107} & \cellcolor{gray!20}\bftab{0.9360$\pm$0.0114} \\
        ~ & DKAJ & 0.1248$\pm$0.0084 & 0.0415$\pm$0.0055 & \cellcolor{gray!20}0.8413$\pm$0.0105 & \cellcolor{gray!20}0.9039$\pm$0.0170 \\
        ~ & DKAJ + SFT & 0.1241$\pm$0.0075 & 0.0409$\pm$0.0051 & \cellcolor{gray!20}0.8415$\pm$0.0102 & \cellcolor{gray!20}0.9037$\pm$0.0181 \\
        \midrule
        \multirow{9}{*}{\rotatebox{90}{Framingham}} & Fine-Gray & 0.0819$\pm$0.0037 & \bftab{0.0566$\pm$0.0035} & \cellcolor{gray!20}\blue{0.7733$\pm$0.0106} & \cellcolor{gray!20}\blue{0.7144$\pm$0.0182} \\
        ~ & cs-Cox & \bftab{0.0814$\pm$0.0039} & \bftab{0.0566$\pm$0.0034} & \cellcolor{gray!20}\bftab{0.7753$\pm$0.0105} & \cellcolor{gray!20}\bftab{0.7160$\pm$0.0173} \\
        ~ & RSF-CR & \blue{0.0818$\pm$0.0032} & 0.0574$\pm$0.0036 & \cellcolor{gray!20}0.7720$\pm$0.0098 & \cellcolor{gray!20}0.6982$\pm$0.0119 \\
        ~ & DeepHit & 0.1168$\pm$0.0034 & 0.0880$\pm$0.0018 & \cellcolor{gray!20}0.7423$\pm$0.0198 & \cellcolor{gray!20}0.6957$\pm$0.0195 \\
        ~ & DSM & 0.0890$\pm$0.0038 & 0.0689$\pm$0.0053 & \cellcolor{gray!20}0.7664$\pm$0.0170 & \cellcolor{gray!20}0.7095$\pm$0.0184 \\
        ~ & NeuralFG & 0.0992$\pm$0.0140 & 0.0628$\pm$0.0037 & \cellcolor{gray!20}0.6833$\pm$0.0619 & \cellcolor{gray!20}0.6978$\pm$0.0287 \\
        ~ & SurvivalBoost & 0.0830$\pm$0.0046 & 0.0571$\pm$0.0036 & \cellcolor{gray!20}0.7645$\pm$0.0152 & \cellcolor{gray!20}0.7031$\pm$0.0156 \\
        ~ & DKAJ & 0.0899$\pm$0.0047 & 0.0579$\pm$0.0038 & \cellcolor{gray!20}0.7677$\pm$0.0138 & \cellcolor{gray!20}0.7076$\pm$0.0209 \\
        ~ & DKAJ + SFT & 0.0928$\pm$0.0108 & 0.0588$\pm$0.0057 & \cellcolor{gray!20}0.7681$\pm$0.0151 & \cellcolor{gray!20}0.7088$\pm$0.0192 \\
        \midrule
        \multirow{9}{*}{\rotatebox{90}{SEER}} & Fine-Gray & 0.0637$\pm$0.0013 & 0.0165$\pm$0.0010 & \cellcolor{gray!20}0.8151$\pm$0.0021 & \cellcolor{gray!20}0.8478$\pm$0.0099 \\
        ~ & cs-Cox & 0.0638$\pm$0.0013 & \bftab{0.0164$\pm$0.0009} & \cellcolor{gray!20}0.7630$\pm$0.0085 & \cellcolor{gray!20}0.8415$\pm$0.0132 \\
        ~ & RSF-CR & \blue{0.0607$\pm$0.0016} & 0.0168$\pm$0.0010 & \cellcolor{gray!20}0.8350$\pm$0.0030 & \cellcolor{gray!20}0.8301$\pm$0.0137 \\
        ~ & DeepHit & 0.0914$\pm$0.0077 & 0.0550$\pm$0.0175 & \cellcolor{gray!20}0.8239$\pm$0.0028 & \cellcolor{gray!20}0.8548$\pm$0.0120 \\
        ~ & DSM & 0.0626$\pm$0.0014 & 0.0186$\pm$0.0019 & \cellcolor{gray!20}0.7807$\pm$0.0088 & \cellcolor{gray!20}0.8422$\pm$0.0119 \\
        ~ & NeuralFG & 0.0611$\pm$0.0016 & 0.0166$\pm$0.0009 & \cellcolor{gray!20}0.7743$\pm$0.0090 & \cellcolor{gray!20}0.8362$\pm$0.0138 \\
        ~ & SurvivalBoost & \bftab{0.0587$\pm$0.0014} & \bftab{0.0164$\pm$0.0009} & \cellcolor{gray!20}\bftab{0.8418$\pm$0.0038} & \cellcolor{gray!20}0.8583$\pm$0.0081 \\
        ~ & DKAJ & 0.0691$\pm$0.0048 & 0.0169$\pm$0.0010 & \cellcolor{gray!20}0.8333$\pm$0.0034 & \cellcolor{gray!20}\bftab{0.8598$\pm$0.0072} \\
        ~ & DKAJ + SFT & 0.0868$\pm$0.0181 & 0.0216$\pm$0.0045 & \cellcolor{gray!20}\blue{0.8371$\pm$0.0043} & \cellcolor{gray!20}\blue{0.8597$\pm$0.0070} \\
        \midrule
        \multirow{9}{*}{\rotatebox{90}{Synthetic}} & Fine-Gray & 0.1851$\pm$0.0035 & 0.1827$\pm$0.0027 & \cellcolor{gray!20}0.5823$\pm$0.0051 & \cellcolor{gray!20}0.5917$\pm$0.0071 \\
        ~ & cs-Cox & 0.1851$\pm$0.0035 & 0.1827$\pm$0.0027 & \cellcolor{gray!20}0.5809$\pm$0.0051 & \cellcolor{gray!20}0.5902$\pm$0.0073 \\
        ~ & RSF-CR & 0.1752$\pm$0.0035 & 0.1742$\pm$0.0027 & \cellcolor{gray!20}0.7224$\pm$0.0071 & \cellcolor{gray!20}0.7203$\pm$0.0043 \\
        ~ & DeepHit & 0.2307$\pm$0.0071 & 0.2265$\pm$0.0088 & \cellcolor{gray!20}\blue{0.7401$\pm$0.0067} & \cellcolor{gray!20}0.7437$\pm$0.0043 \\
        ~ & DSM & 0.1750$\pm$0.0036 & 0.1725$\pm$0.0035 & \cellcolor{gray!20}0.7280$\pm$0.0055 & \cellcolor{gray!20}0.7320$\pm$0.0042 \\
        ~ & NeuralFG & \bftab{0.1634$\pm$0.0039} & \bftab{0.1612$\pm$0.0031} & \cellcolor{gray!20}\bftab{0.7481$\pm$0.0073} & \cellcolor{gray!20}\bftab{0.7513$\pm$0.0045} \\
        ~ & SurvivalBoost & 0.1753$\pm$0.0033 & 0.1736$\pm$0.0026 & \cellcolor{gray!20}0.7160$\pm$0.0083 & \cellcolor{gray!20}0.7190$\pm$0.0039 \\
        ~ & DKAJ & \blue{0.1729$\pm$0.0048} & \blue{0.1700$\pm$0.0040} & \cellcolor{gray!20}0.7384$\pm$0.0067 & \cellcolor{gray!20}0.7435$\pm$0.0041 \\
        ~ & DKAJ + SFT & 0.1783$\pm$0.0033 & 0.1757$\pm$0.0031 & \cellcolor{gray!20}0.7388$\pm$0.0072 & \cellcolor{gray!20}\blue{0.7439$\pm$0.0044} \\
        \bottomrule
    \end{tabular}
    \end{adjustbox}
    }
\end{table*}

\subsection{Pairwise Performance Significance}
\label{apd:pairwise-tests}

To assess the statistical significance of performance differences among models, we conduct paired, two-sided Wilcoxon signed-rank tests for both metrics: IBS and \Ctd.
Each test pools 80 paired observations ($4~\text{datasets} \times 10~\text{random splits} \times 2~\text{events}$), with pairs matched on dataset, split, and event to ensure comparable conditions across methods.

To align evaluation with the selection objective, we run two independent analyses:
(i) for the IBS tests, model instances are chosen based on the best validation IBS;
(ii) for the Ctd tests, model instances are chosen based on the best validation Ctd.
As discussed in the main text, the validation criterion can influence relative rankings, so separating the analyses avoids conflating selection effects across metrics.

For each of the evaluation metrics (IBS or \Ctd), we conduct the test separately using the following steps:
\begin{enumerate}[noitemsep,topsep=2pt]
    \item For each unordered method pair and for the relevant metric/selection setting, we compute paired differences across the 80 matched observations (ignoring ties).
    \item We apply the Wilcoxon signed-rank test (two-sided) to obtain one p-value per unordered pair of methods.
    \item To control the family-wise error rate within each metric, we apply the Holm–Bonferroni correction across all unordered pairs, yielding adjusted p-values.
\end{enumerate}

The complete pairwise outcomes are provided in Tables~\ref{tab:pairwise-ibs} (IBS, models selected by IBS) and~\ref{tab:pairwise-ctd} (\Ctd, models selected by \Ctd).
Across metrics, DKAJ is broadly competitive with the strongest baselines and exhibits statistically significant improvements over several classical and deep models in numerous pairings.
Differences among top-performing neural and ensemble methods (e.g., SurvivalBoost and NeuralFG) are often not statistically significant after Holm correction, indicating comparable performance levels within this group.
These results reinforce the aggregate findings presented in the main paper and highlight the stability of DKAJ’s performance across both discrimination and calibration metrics under consistent statistical testing.

\begin{table*}[t!]
\centering
    \floatconts {tab:pairwise-ibs}
    {\caption{Paired, two-sided Wilcoxon signed-rank tests on \textbf{IBS} (lower is better), 
    pooled over 4 datasets $\times$ 10 splits $\times$ 2 events (80 pairs per comparison; 
    pairs matched on dataset, split, and event). Cells show the row–vs–column 
    direction and Holm–Bonferroni–corrected significance across all unordered 
    method pairs for IBS: $\downarrow$ = row lower (better), $\uparrow$ = row higher (worse); 
    ***: \(\tilde p<0.001\), **: \(\tilde p<0.01\), *: \(\tilde p<0.05\), 
    \(\approx\): not significant, where \(\tilde p\) denotes the Holm-adjusted $p$-value.} \vspace{-1.5em}}
    {
    \begin{adjustbox}{max width=0.7\textwidth}
    \begin{tabular}{llllllllll}
    \toprule
    {} & \rotatebox{80}{Fine-Gray} & \rotatebox{80}{cs-Cox} & \rotatebox{80}{RSF-CR} & \rotatebox{80}{DeepHit} & \rotatebox{80}{DSM} & \rotatebox{80}{NeuralFG} & \rotatebox{80}{SurvivalBoost} & \rotatebox{80}{DKAJ} & \rotatebox{80}{DKAJ+SFT} \\
    \midrule
    Fine-Gray     &         - &    $\uparrow^{**}$ &   $\uparrow^{***}$ &       $\approx$ &     $\approx$ &        $\approx$ &          $\uparrow^{***}$ &   $\uparrow^{**}$ &        $\approx$ \\
    cs-Cox        &       $\downarrow^{**}$ &      — &   $\uparrow^{***}$ &      $\downarrow^{*}$ &     $\approx$ &        $\approx$ &          $\uparrow^{***}$ &    $\uparrow^{*}$ &        $\approx$ \\
    RSF-CR        &      $\downarrow^{***}$ &   $\downarrow^{***}$ &      — &    $\downarrow^{***}$ &  $\downarrow^{***}$ &        $\approx$ &          $\uparrow^{***}$ &     $\approx$ &      $\downarrow^{**}$ \\
    DeepHit       &         $\approx$ &     $\uparrow^{*}$ &   $\uparrow^{***}$ &       — &  $\uparrow^{***}$ &     $\uparrow^{***}$ &          $\uparrow^{***}$ &  $\uparrow^{***}$ &      $\uparrow^{**}$ \\
    DSM           &         $\approx$ &      $\approx$ &   $\uparrow^{***}$ &    $\downarrow^{***}$ &     — &     $\uparrow^{***}$ &          $\uparrow^{***}$ &  $\uparrow^{***}$ &        $\approx$ \\
    NeuralFG      &         $\approx$ &      $\approx$ &      $\approx$ &    $\downarrow^{***}$ &  $\downarrow^{***}$ &        — &             $\approx$ &     $\approx$ &       $\downarrow^{*}$ \\
    SurvivalBoost &      $\downarrow^{***}$ &   $\downarrow^{***}$ &   $\downarrow^{***}$ &    $\downarrow^{***}$ &  $\downarrow^{***}$ &        $\approx$ &             — &     $\approx$ &     $\downarrow^{***}$ \\
    DKAJ          &       $\downarrow^{**}$ &     $\downarrow^{*}$ &      $\approx$ &    $\downarrow^{***}$ &  $\downarrow^{***}$ &        $\approx$ &             $\approx$ &     — &     $\downarrow^{***}$ \\
    DKAJ+SFT      &         $\approx$ &      $\approx$ &    $\uparrow^{**}$ &     $\downarrow^{**}$ &     $\approx$ &       $\uparrow^{*}$ &          $\uparrow^{***}$ &  $\uparrow^{***}$ &        — \\
    \bottomrule
    \end{tabular}
    \end{adjustbox}
    }
\end{table*}

\begin{table*}[t!]
\centering
    \floatconts {tab:pairwise-ctd}
    {\caption{Paired, two-sided Wilcoxon signed-rank tests on \textbf{\Ctd} (higher is better), 
    pooled over 4 datasets $\times$ 10 splits $\times$ 2 events (80 pairs per comparison; 
    pairs matched on dataset, split, and event). Cells show the row–vs–column 
    direction and Holm–Bonferroni–corrected significance across all unordered 
    method pairs for \Ctd: $\uparrow$ = row higher (better), $\downarrow$ = row lower (worse); 
    ***: \(\tilde p<0.001\), **: \(\tilde p<0.01\), *: \(\tilde p<0.05\), 
    \(\approx\): not significant, where \(\tilde p\) denotes the Holm-adjusted $p$-value.} \vspace{-1.5em}}
    {
    \begin{adjustbox}{max width=0.7\textwidth}
    \begin{tabular}{llllllllll}
    \toprule
    {} & \rotatebox{80}{Fine-Gray} & \rotatebox{80}{cs-Cox} & \rotatebox{80}{RSF-CR} & \rotatebox{80}{DeepHit} & \rotatebox{80}{DSM} & \rotatebox{80}{NeuralFG} & \rotatebox{80}{SurvivalBoost} & \rotatebox{80}{DKAJ} & \rotatebox{80}{DKAJ+SFT} \\
    \midrule
    Fine-Gray     &         — &      $\approx$ &    $\downarrow^{**}$ &     $\downarrow^{**}$ &     $\approx$ &        $\approx$ &          $\downarrow^{***}$ &  $\downarrow^{***}$ &     $\downarrow^{***}$ \\
    cs-Cox        &         $\approx$ &      — &    $\downarrow^{**}$ &     $\downarrow^{**}$ &   $\downarrow^{**}$ &        $\approx$ &          $\downarrow^{***}$ &  $\downarrow^{***}$ &     $\downarrow^{***}$ \\
    RSF-CR        &       $\uparrow^{**}$ &    $\uparrow^{**}$ &      — &       $\approx$ &     $\approx$ &        $\approx$ &           $\downarrow^{**}$ &  $\downarrow^{***}$ &     $\downarrow^{***}$ \\
    DeepHit       &       $\uparrow^{**}$ &    $\uparrow^{**}$ &      $\approx$ &       — &     $\approx$ &        $\approx$ &             $\approx$ &    $\downarrow^{*}$ &      $\downarrow^{**}$ \\
    DSM           &         $\approx$ &    $\uparrow^{**}$ &      $\approx$ &       $\approx$ &     — &        $\approx$ &           $\downarrow^{**}$ &  $\downarrow^{***}$ &     $\downarrow^{***}$ \\
    NeuralFG      &         $\approx$ &      $\approx$ &      $\approx$ &       $\approx$ &     $\approx$ &        — &           $\downarrow^{**}$ &   $\downarrow^{**}$ &      $\downarrow^{**}$ \\
    SurvivalBoost &      $\uparrow^{***}$ &   $\uparrow^{***}$ &    $\uparrow^{**}$ &       $\approx$ &   $\uparrow^{**}$ &      $\uparrow^{**}$ &             — &     $\approx$ &        $\approx$ \\
    DKAJ          &      $\uparrow^{***}$ &   $\uparrow^{***}$ &   $\uparrow^{***}$ &      $\uparrow^{*}$ &  $\uparrow^{***}$ &      $\uparrow^{**}$ &             $\approx$ &     — &        $\approx$ \\
    DKAJ+SFT      &      $\uparrow^{***}$ &   $\uparrow^{***}$ &   $\uparrow^{***}$ &     $\uparrow^{**}$ &  $\uparrow^{***}$ &      $\uparrow^{**}$ &             $\approx$ &     $\approx$ &        — \\
    \bottomrule
    \end{tabular}
    \end{adjustbox}
    }
\end{table*}

\subsection{Performance Using Alternative Early Stopping} \label{apd:alt-es}
As described in Appendix~\ref{apd:exp-hyperparam}, the deep learning–based methods differ slightly in their early stopping criteria. Specifically, DeepHit and DKAJ employ \emph{metric-aligned early stopping}, where training stops once the validation performance metric (IBS or~$C^{\text{td}}$) plateaus for 10 epochs. In contrast, DSM and NeuralFG follow \emph{objective-based early stopping}, terminating training when the learning objective on the validation set ceases to improve for 10 epochs. Although both approaches are standard, this introduces a minor inconsistency across models in how the stopping signal is defined.

To assess the sensitivity of our results to the stopping criterion, we retrained DeepHit and DKAJ using the same objective-based stopping rule as DSM and NeuralFG (validation objective, patience~10). For these runs, model checkpoints were saved according to the best validation objective, but the final model selected for reporting test performance, among all hyperparameter configurations, remained based on the best validation metric (IBS or~$C^{\text{td}}$), consistent with our main protocol. 
Tables~\ref{tab:early-stop-by-val-loss-by-ibs} and~\ref{tab:early-stop-by-val-loss-by-ctd} compare the two stopping strategies. Across datasets and events, the differences are numerically negligible and do not alter conclusions. We therefore retain metric-aligned early stopping for DeepHit and DKAJ in the main experiments.

\begin{table*}[!ht]
    \centering
    \floatconts {tab:early-stop-by-val-loss-by-ibs}
    {\caption{Comparison of metric-aligned vs.~objective-based early stopping when models are selected by best validation IBS. 
    For DeepHit and DKAJ, ``(ObjStop)'' denotes runs stopped based on the validation objective (patience~10), as used for DSM and NeuralFG. 
    Model selection for reporting test results remains based on best validation IBS; the IBS columns are shaded to highlight the results. Differences due to the stopping rule are negligible.}\vspace{-1em}}
    {
    \begin{adjustbox}{max width=0.85\textwidth}
    \begin{tabular}{llcccc}
        \toprule
        \bfseries \multirow{2}{*}{Dataset} & \bfseries \multirow{2}{*}{Method} & \multicolumn{2}{c}{\cellcolor{gray!20}\bfseries IBS $(\downarrow)$}  & \multicolumn{2}{c}{\bfseries $C^{\text{td}}$ $(\uparrow)$} \\
        ~ & ~ & \cellcolor{gray!20}Primary & \cellcolor{gray!20}Competing & Primary & Competing \\
        \midrule
        \multirow{4}{*}{PBC} & DeepHit & \cellcolor{gray!20}0.1259$\pm$0.0121 & \cellcolor{gray!20}0.0552$\pm$0.0074 & 0.8243$\pm$0.0275 & 0.8050$\pm$0.0765 \\
        ~ & DeepHit (ObjStop) & \cellcolor{gray!20}0.1337$\pm$0.0099 & \cellcolor{gray!20}0.0652$\pm$0.0079 & 0.8280$\pm$0.0154 & 0.8035$\pm$0.0757 \\
        ~ & DKAJ & \cellcolor{gray!20}0.1031$\pm$0.0071 & \cellcolor{gray!20}0.0394$\pm$0.0057 & 0.8242$\pm$0.0215 & 0.7885$\pm$0.0825 \\
        ~ & DKAJ (ObjStop) & \cellcolor{gray!20}0.1093$\pm$0.0080 & \cellcolor{gray!20}0.0401$\pm$0.0056 & 0.8043$\pm$0.0466 & 0.7996$\pm$0.1260 \\
        \midrule
        \multirow{4}{*}{Framingham} & DeepHit & \cellcolor{gray!20}0.0941$\pm$0.0049 & \cellcolor{gray!20}0.0645$\pm$0.0055 & 0.7099$\pm$0.0146 & 0.6464$\pm$0.0224 \\
        ~ & DeepHit (ObjStop) & \cellcolor{gray!20}0.1106$\pm$0.0080 & \cellcolor{gray!20}0.0669$\pm$0.0076 & 0.7006$\pm$0.0187 & 0.6549$\pm$0.0238 \\
        ~ & DKAJ & \cellcolor{gray!20}0.0844$\pm$0.0036 & \cellcolor{gray!20}0.0578$\pm$0.0034 & 0.7613$\pm$0.0155 & 0.6870$\pm$0.0192 \\
        ~ & DKAJ (ObjStop) & \cellcolor{gray!20}0.0836$\pm$0.0031 & \cellcolor{gray!20}0.0578$\pm$0.0034 & 0.7651$\pm$0.0148 & 0.6867$\pm$0.0326 \\
        \midrule
        \multirow{4}{*}{SEER} & DeepHit & \cellcolor{gray!20}0.0721$\pm$0.0012 & \cellcolor{gray!20}0.0228$\pm$0.0017 & 0.8043$\pm$0.0140 & 0.8134$\pm$0.0494 \\
        ~ & DeepHit (ObjStop) & \cellcolor{gray!20}0.0753$\pm$0.0022 & \cellcolor{gray!20}0.0237$\pm$0.0020 & 0.8116$\pm$0.0179 & 0.8435$\pm$0.0125 \\
        ~ & DKAJ & \cellcolor{gray!20}0.0609$\pm$0.0011 & \cellcolor{gray!20}0.0167$\pm$0.0011 & 0.8277$\pm$0.0043 & 0.8250$\pm$0.0153 \\
        ~ & DKAJ (ObjStop) & \cellcolor{gray!20}0.0612$\pm$0.0013 & \cellcolor{gray!20}0.0167$\pm$0.0010 & 0.8297$\pm$0.0054 & 0.8354$\pm$0.0150 \\
        \midrule
        \multirow{4}{*}{Synthetic} & DeepHit & \cellcolor{gray!20}0.1732$\pm$0.0046 & \cellcolor{gray!20}0.1702$\pm$0.0026 & 0.7214$\pm$0.0181 & 0.7233$\pm$0.0127 \\
        ~ & DeepHit (ObjStop) & \cellcolor{gray!20}0.1733$\pm$0.0046 & \cellcolor{gray!20}0.1695$\pm$0.0027 & 0.7334$\pm$0.0073 & 0.7376$\pm$0.0063 \\
        ~ & DKAJ & \cellcolor{gray!20}0.1672$\pm$0.0039 & \cellcolor{gray!20}0.1647$\pm$0.0029 & 0.7318$\pm$0.0083 & 0.7346$\pm$0.0052 \\
        ~ & DKAJ (ObjStop) & \cellcolor{gray!20}0.1667$\pm$0.0037 & \cellcolor{gray!20}0.1640$\pm$0.0026 & 0.7320$\pm$0.0068 & 0.7367$\pm$0.0056 \\
        \bottomrule
    \end{tabular}
    \end{adjustbox}
    }
\end{table*}

\begin{table*}[!ht]
    \centering
    \floatconts {tab:early-stop-by-val-loss-by-ctd}
    {\caption{Comparison of metric-aligned vs.~objective-based early stopping when models are selected by best validation \Ctd. 
    For DeepHit and DKAJ, ``(ObjStop)'' denotes runs stopped based on the validation objective (patience~10), as used for DSM and NeuralFG. 
    Model selection for reporting test results remains based on best validation \Ctd; the \Ctd~columns are shaded to highlight the results. Differences due to the stopping rule are negligible.}\vspace{-1em}}
    {
    \begin{adjustbox}{max width=0.85\textwidth}
    \begin{tabular}{llcccc}
        \toprule
        \bfseries \multirow{2}{*}{Dataset} & \bfseries \multirow{2}{*}{Method} & \multicolumn{2}{c}{\bfseries IBS $(\downarrow)$}  & \multicolumn{2}{c}{\cellcolor{gray!20}\bfseries $C^{\text{td}}$ $(\uparrow)$} \\
        ~ & ~ & Primary & Competing & \cellcolor{gray!20}Primary & \cellcolor{gray!20}Competing \\
        \midrule
        \multirow{4}{*}{PBC} & DeepHit & 0.1308$\pm$0.0052 & 0.0785$\pm$0.0086 & \cellcolor{gray!20}0.8407$\pm$0.0131 & \cellcolor{gray!20}0.9060$\pm$0.0130 \\
        ~ & DeepHit (ObjStop) & 0.1343$\pm$0.0062 & 0.0892$\pm$0.0127 & \cellcolor{gray!20}0.8322$\pm$0.0145 & \cellcolor{gray!20}0.8946$\pm$0.0185 \\
        ~ & DKAJ & 0.1248$\pm$0.0084 & 0.0415$\pm$0.0055 & \cellcolor{gray!20}0.8413$\pm$0.0105 & \cellcolor{gray!20}0.9039$\pm$0.0170 \\
        ~ & DKAJ (ObjStop) & 0.1240$\pm$0.0073 & 0.0417$\pm$0.0058 & \cellcolor{gray!20}0.8406$\pm$0.0077 & \cellcolor{gray!20}0.9044$\pm$0.0140 \\
        \midrule
        \multirow{4}{*}{Framingham} & DeepHit & 0.1168$\pm$0.0034 & 0.0880$\pm$0.0018 & \cellcolor{gray!20}0.7423$\pm$0.0198 & \cellcolor{gray!20}0.6957$\pm$0.0195 \\
        ~ & DeepHit (ObjStop) & 0.1153$\pm$0.0045 & 0.0848$\pm$0.0085 & \cellcolor{gray!20}0.7423$\pm$0.0178 & \cellcolor{gray!20}0.6907$\pm$0.0212 \\
        ~ & DKAJ & 0.0899$\pm$0.0047 & 0.0579$\pm$0.0038 & \cellcolor{gray!20}0.7677$\pm$0.0138 & \cellcolor{gray!20}0.7076$\pm$0.0209 \\
        ~ & DKAJ (ObjStop) & 0.0900$\pm$0.0048 & 0.0579$\pm$0.0037 & \cellcolor{gray!20}0.7673$\pm$0.0119 & \cellcolor{gray!20}0.7112$\pm$0.0192 \\
        \midrule
        \multirow{4}{*}{SEER} & DeepHit & 0.0914$\pm$0.0077 & 0.0550$\pm$0.0175 & \cellcolor{gray!20}0.8239$\pm$0.0028 & \cellcolor{gray!20}0.8548$\pm$0.0120 \\
        ~ & DeepHit (ObjStop) & 0.0863$\pm$0.0089 & 0.0483$\pm$0.0162 & \cellcolor{gray!20}0.8244$\pm$0.0042 & \cellcolor{gray!20}0.8548$\pm$0.0095 \\
        ~ & DKAJ & 0.0691$\pm$0.0048 & 0.0169$\pm$0.0010 & \cellcolor{gray!20}0.8333$\pm$0.0034 & \cellcolor{gray!20}0.8598$\pm$0.0072 \\
        ~ & DKAJ (ObjStop) & 0.0692$\pm$0.0048 & 0.0169$\pm$0.0010 & \cellcolor{gray!20}0.8334$\pm$0.0045 & \cellcolor{gray!20}0.8566$\pm$0.0113 \\
        \midrule
        \multirow{4}{*}{Synthetic} & DeepHit & 0.2307$\pm$0.0071 & 0.2265$\pm$0.0088 & \cellcolor{gray!20}0.7401$\pm$0.0067 & \cellcolor{gray!20}0.7437$\pm$0.0043 \\
        ~ & DeepHit (ObjStop) & 0.2299$\pm$0.0129 & 0.2239$\pm$0.0088 & \cellcolor{gray!20}0.7408$\pm$0.0072 & \cellcolor{gray!20}0.7439$\pm$0.0061 \\
        ~ & DKAJ & 0.1729$\pm$0.0048 & 0.1700$\pm$0.0040 & \cellcolor{gray!20}0.7384$\pm$0.0067 & \cellcolor{gray!20}0.7435$\pm$0.0041 \\
        ~ & DKAJ (ObjStop) & 0.1744$\pm$0.0037 & 0.1722$\pm$0.0027 & \cellcolor{gray!20}0.7398$\pm$0.0067 & \cellcolor{gray!20}0.7445$\pm$0.0044 \\
        \bottomrule
    \end{tabular}
    \end{adjustbox}
    }
\end{table*}

\subsection{Ablation Study} \label{apd:ablation}

We ablate two components of DKAJ to assess their necessity: (i) the leave-one-out (LOO) likelihood in \eqref{eq:DKAJ-log-likelihood-train}, replaced by the standard likelihood in \eqref{eq:DKAJ-log-likelihood} to form \textit{noLOO}; and (ii) the TUNA warm start, replaced by PyTorch’s default initialization to form \mbox{\textit{noTUNA}}. For each ablation we toggle only the targeted component and keep the architecture, hyperparameter grids, early-stopping policy, and search protocol identical to DKAJ (see Appendix~\ref{apd:exp-hyperparam} for details on grids and optimization).

To avoid mixing selection criteria across metrics, we use metric-aligned model selection: IBS values are reported from the model chosen by the best validation IBS, and $C^{\text{td}}$ values are reported from the model chosen by the best validation \Ctd. We report results in terms of mean~$\pm$~standard deviation over 10 random splits.

When training without the LOO formulation, validation performance typically deteriorates early, indicating overfitting—consistent with the intended regularizing role of excluding each subject’s own label from its kernel-weighted hazards. The TUNA warm start, by contrast, primarily stabilizes optimization and improves the initial embedding geometry. As noted in Appendix~\ref{apd:exp-hyperparam}, this two-stage scheme also reduces the effective hyperparameter search space during the main training. %

Table~\ref{tab:ablation-results} summarizes the ablations on three real-world datasets (PBC, Framingham, SEER) and one synthetic dataset.
On the real-world datasets, removing either component generally worsens performance in most cases. In particular, noLOO shows lower \Ctd~and higher IBS on SEER and Framingham, and exhibits increased variability on PBC, while noTUNA yields smaller changes relative to DKAJ. On the synthetic dataset, noLOO substantially degrades both metrics, whereas noTUNA is comparable to (and in some entries slightly better than) DKAJ. Overall, these results indicate that both components contribute to the robustness of DKAJ.

\begin{table*}[!ht]
    \centering
    \floatconts {tab:ablation-results}
    {\caption{Ablation results with metric-aligned reporting: IBS is selected by best validation IBS; \Ctd~is selected by best validation \Ctd. Mean~$\pm$~std.~dev.~over 10 splits.}\vspace{-1em}}
    {
    \begin{adjustbox}{max width=0.8\textwidth}
    \begin{tabular}{cccccc}
        \toprule
        \bfseries \multirow{2}{*}{Dataset} & \bfseries \multirow{2}{*}{Method} & \multicolumn{2}{c}{\bfseries IBS $(\downarrow)$}  & \multicolumn{2}{c}{\bfseries $C^{\text{td}}$ $(\uparrow)$} \\
        ~ & ~ & Primary & Competing & Primary & Competing \\
        \midrule
        \multirow{3}{*}{PBC} & DKAJ & 0.1031$\pm$0.0071 & 0.0394$\pm$0.0057 & \bftab{0.8413$\pm$0.0105} & \bftab{0.9039$\pm$0.0170} \\
        ~ & DKAJ noLOO & \bftab{0.1012$\pm$0.0077} & \bftab{0.0393$\pm$0.0056} & 0.8382$\pm$0.0137 & 0.8827$\pm$0.0371 \\
        ~ & DKAJ noTUNA & 0.1049$\pm$0.0054 & 0.0399$\pm$0.0054 & 0.8399$\pm$0.0114 & 0.8977$\pm$0.0258 \\
        \midrule
        \multirow{3}{*}{Framingham} & DKAJ & \bftab{0.0844$\pm$0.0036} & \bftab{0.0578$\pm$0.0034} & \bftab{0.7677$\pm$0.0138} & \bftab{0.7076$\pm$0.0209} \\
        ~ & DKAJ noLOO & 0.0869$\pm$0.0056 & 0.0579$\pm$0.0035 & 0.7644$\pm$0.0131 & 0.6973$\pm$0.0203 \\
        ~ & DKAJ noTUNA & 0.0848$\pm$0.0030 & \bftab{0.0578$\pm$0.0037} & 0.7644$\pm$0.0149 & 0.7041$\pm$0.0191 \\
        \midrule
        \multirow{3}{*}{SEER} & DKAJ & \bftab{0.0609$\pm$0.0011} & \bftab{0.0167$\pm$0.0011} & \bftab{0.8333$\pm$0.0034} & \bftab{0.8598$\pm$0.0072} \\
        ~ & DKAJ noLOO & 0.0637$\pm$0.0019 & 0.0170$\pm$0.0009 & 0.8175$\pm$0.0068 & 0.8313$\pm$0.0129 \\
        ~ & DKAJ noTUNA & 0.0617$\pm$0.0013 & 0.0169$\pm$0.0010 & 0.8256$\pm$0.0040 & 0.8585$\pm$0.0076 \\
        \midrule
        \multirow{3}{*}{Synthetic} & DKAJ & 0.1672$\pm$0.0039 & 0.1647$\pm$0.0029 & 0.7384$\pm$0.0067 & 0.7435$\pm$0.0041 \\
        ~ & DKAJ noLOO & 0.1808$\pm$0.0033 & 0.1790$\pm$0.0029 & 0.6174$\pm$0.0098 & 0.6279$\pm$0.0098 \\
        ~ & DKAJ noTUNA & \bftab{0.1660$\pm$0.0037} & \bftab{0.1637$\pm$0.0028} & \bftab{0.7427$\pm$0.0071} & \bftab{0.7461$\pm$0.0048} \\
        \bottomrule
    \end{tabular}
    \end{adjustbox}
    }
\end{table*}

\subsection{Varying Training Size} \label{apd:varying-training-size}
To examine robustness under different data availability, we vary the proportion of data used for training from 10\% to 70\%, reserving the remaining 30\% as a held-out test set.
Each configuration includes both training and validation data within the specified proportion.
Results are averaged over 10 random splits.

Figures \ref{fig:C-td-varying-training-size-e1} and \ref{fig:C-td-varying-training-size-e2} show the $C^{\text{td}}$ for the two competing events.
Across methods, performance generally improves as the training size increases and stabilizes beyond approximately 40–50\%.
DKAJ scales smoothly with data size and follows similar performance trends to other neural-network-based models (DeepHit, NeuralFG) and the ensemble baseline (SurvivalBoost).
We observe no indication of overfitting even in smaller training regimes, confirming that DKAJ maintains stable and data-efficient behavior across varying cohort sizes.

\begin{figure}[t!]
\floatconts
  {fig:C-td-varying-training-size-e2}
  {\caption{\Ctd~on the synthetic dataset (Event 2) as training set size varies; 30\% held-out test.}\vspace{-1.1em}}
  {\vspace{-.5em}\includegraphics[width=.95\linewidth]{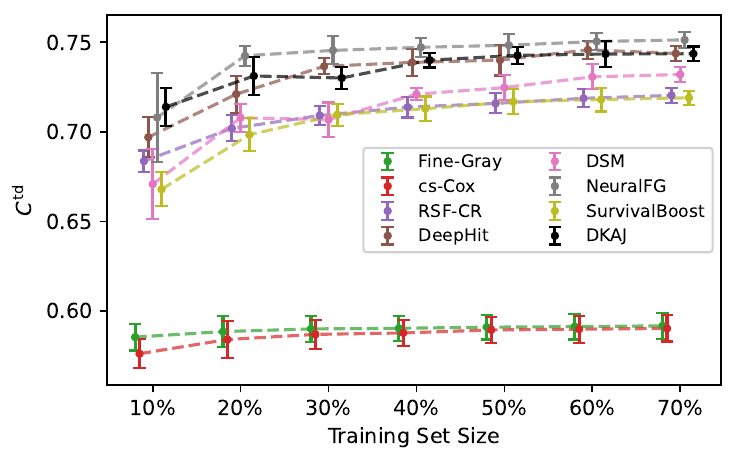}\vspace{-3em}}
\end{figure}

\subsection{Computation Cost} \label{apd:compute-cost}

We report wall–clock training time and model size on the Framingham dataset. Times are averaged over 10 random splits (mean~$\pm$~std), and reflect the \emph{total} time per split including the full hyperparameter sweep under IBS–based model selection. Model size is summarized as the number of learnable neural–network parameters; for DKAJ, we also report the number of entries in the (deterministic) summary tables used at inference (not trainable and not part of backprop).

\paragraph{Training time}
Table~\ref{tab:training_time} shows the total training time per method. DKAJ is the same order of magnitude as other deep baselines (faster than NeuralFG under our grid), and slower than shallow or classical models. Table~\ref{tab:training_time-decompose} decomposes DKAJ’s time into: SurvivalBoost pretraining (used by the warm start), the TUNA warm–up network fitting, and the main DKAJ training. The warm–up overhead is small; the total is dominated by SurvivalBoost and the main DKAJ phase.

\begin{table}[htbp]
\centering
    \floatconts{tab:training_time}
    {\caption{Average total training time per split (minutes; mean~$\pm$~std over 10 splits) on Framingham. Times include the full hyperparameter sweep.}\vspace{-1em}}
    {\setlength{\tabcolsep}{3pt}
    \begin{adjustbox}{max width=0.27\textwidth}
    \begin{tabular}{lc}
    \toprule
        \textbf{Method} & \textbf{Time (minutes)} \\
        \midrule
        Fine-Gray & $0.064 \pm 0.015$ \\
        cs-Cox & $0.006 \pm 0.000$ \\
        RSF-CR & $0.142 \pm 0.072$ \\
        DeepHit & $1.279 \pm 0.461$ \\
        DSM & $7.894 \pm 1.397$ \\
        NeuralFG & $32.176 \pm 29.290$ \\
        SurvivalBoost & $13.526 \pm 1.333$ \\
        DKAJ (total) & $22.813 \pm 2.433$ \\
    \bottomrule
    \end{tabular}
    \end{adjustbox}
    }
\end{table}

\begin{table}[htbp]
\centering
    \floatconts{tab:training_time-decompose}
    {\vspace{-1em}\caption{DKAJ training time decomposition on Framingham (minutes; mean~$\pm$~std over 10 splits). ``SurvivalBoost'' is used by the warm start; ``TUNA warm–up'' initializes neural net weights to approximate SurvivalBoost prediction; ``DKAJ (main) is the main training using a DeepHit loss function.}\vspace{-1em}}
    {\setlength{\tabcolsep}{3pt}
    \begin{adjustbox}{max width=0.36\textwidth}
    \begin{tabular}{lc}
    \toprule
        \textbf{Method} & \textbf{Time (minutes)} \\
        \midrule
        SurvivalBoost & $13.526 \pm 1.333$ \\
        DKAJ (TUNA warm-up) & $0.295 \pm 0.091$ \\
        DKAJ (main) & $8.992 \pm 1.883$ \\
        DKAJ (total) & $22.813 \pm 2.433$ \\
    \bottomrule
    \end{tabular}
    \end{adjustbox}
    }
\end{table}

\paragraph{Model size and storage}
Table~\ref{tab:model_parameters} reports the number of learnable parameters for each neural baseline and, for DKAJ, the number of entries in the summary tables used for cluster/time lookups at inference. These summary tables scale with the number of clusters and the time grid, but they are fixed (non–trainable) and do not contribute to backprop memory. The trainable parameter count of DKAJ is comparable to other deep baselines.

\begin{table*}[t!]
\centering
    \floatconts{tab:model_parameters}
    {\caption{Model size (learnable parameters) and, for DKAJ, the number of entries in the deterministic summary tables used at inference. Values are mean~$\pm$~std across hyperparameter settings and splits.\vspace{-2em}}}
    {\setlength{\tabcolsep}{3pt}
    \begin{adjustbox}{max width=0.7\textwidth}
    \begin{tabular}{l c c}
    \toprule
        \textbf{Method} & \makecell{\textbf{No. parameters } \\ (neural net)} & \makecell{\textbf{No. entries in DKAJ summary functions}\\ (equation~\eqref{eq:cluster-event-count}, ~\eqref{eq:cluster-at-risk-count})} \\
        \midrule
        DeepHit   & $7581\text{k} \pm 25\text{k}$   & $\sim$ \\
        DSM       & $1560\text{k} \pm 0\text{k}$    & $\sim$ \\
        NeuralFG  & $3216\text{k} \pm 0\text{k}$    & $\sim$ \\
        DKAJ      & $2547\text{k} \pm 957\text{k}$  & $71373\text{k} \pm 10275\text{k}$ \\
    \bottomrule
    \end{tabular}
    \end{adjustbox}
    }
\end{table*}

\section{Additional Visualization that Help with Interpretation} \label{apd:visualization}

\subsection{Individual-Level Visualizations} \label{apd:visualization-individual-level}
As discussed in Sections~\ref{sec:DKAJ-overall} and \ref{sec:experiments}, a trained DKAJ estimator inherently supports individual-level interpretation. We illustrate this using the Framingham dataset, although the same approach applies to any dataset.  

For a given test data point, the DKAJ model produces predicted CIFs for each competing event, along with weights that quantify the contribution of different clusters to the prediction. Figure~\ref{fig:individual-cif-framingham-top-5} displays the CIFs (i.e., Aalen–Johansen curves) of the 5 clusters with the highest weights for a randomly chosen test patient, together with the model’s overall prediction for that individual. This allows us to identify which clusters are most influential in shaping the prediction.  

To better understand the defining characteristics of these influential clusters, Figure~\ref{fig:individual-heatmap-framingham-top-5} shows a feature-level heatmap summarizing the distribution of features across the same 5 clusters. By examining these cluster profiles, we can link predictive behavior to specific clinical factors. 

\begin{figure}[htbp]
\floatconts
  {fig:individual-cif-framingham-top-5}
  {\caption{(Framingham, individual-level) Top 5 clusters with the highest weights assigned to a randomly chosen test patient. We show the AJ curves for the 5 clusters as well as the predicted CIF for this test patient. Clusters correspond across the two plots in this figure as well as in Figure~\ref{fig:individual-heatmap-framingham-top-5}.}}
  {\includegraphics[width=1.0\linewidth]{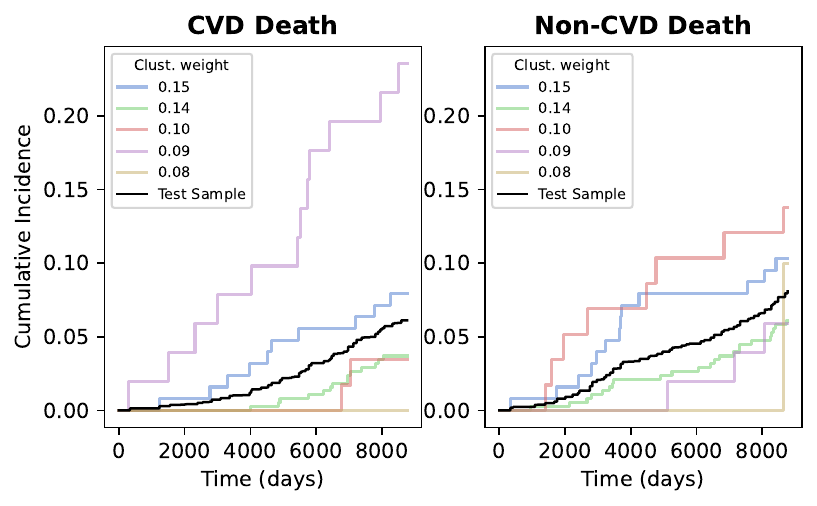}\vspace{-2.5em}}
\end{figure}

\begin{figure}[t!]
\floatconts
  {fig:individual-heatmap-framingham-top-5}
  {\caption{(Framingham, individual-level) Feature heatmap summarizing distributions of variables in the same 5 clusters as in Figure~\ref{fig:individual-cif-framingham-top-5}. Darker shades mean higher feature values or frequencies.}}
  {\includegraphics[width=1.\linewidth]{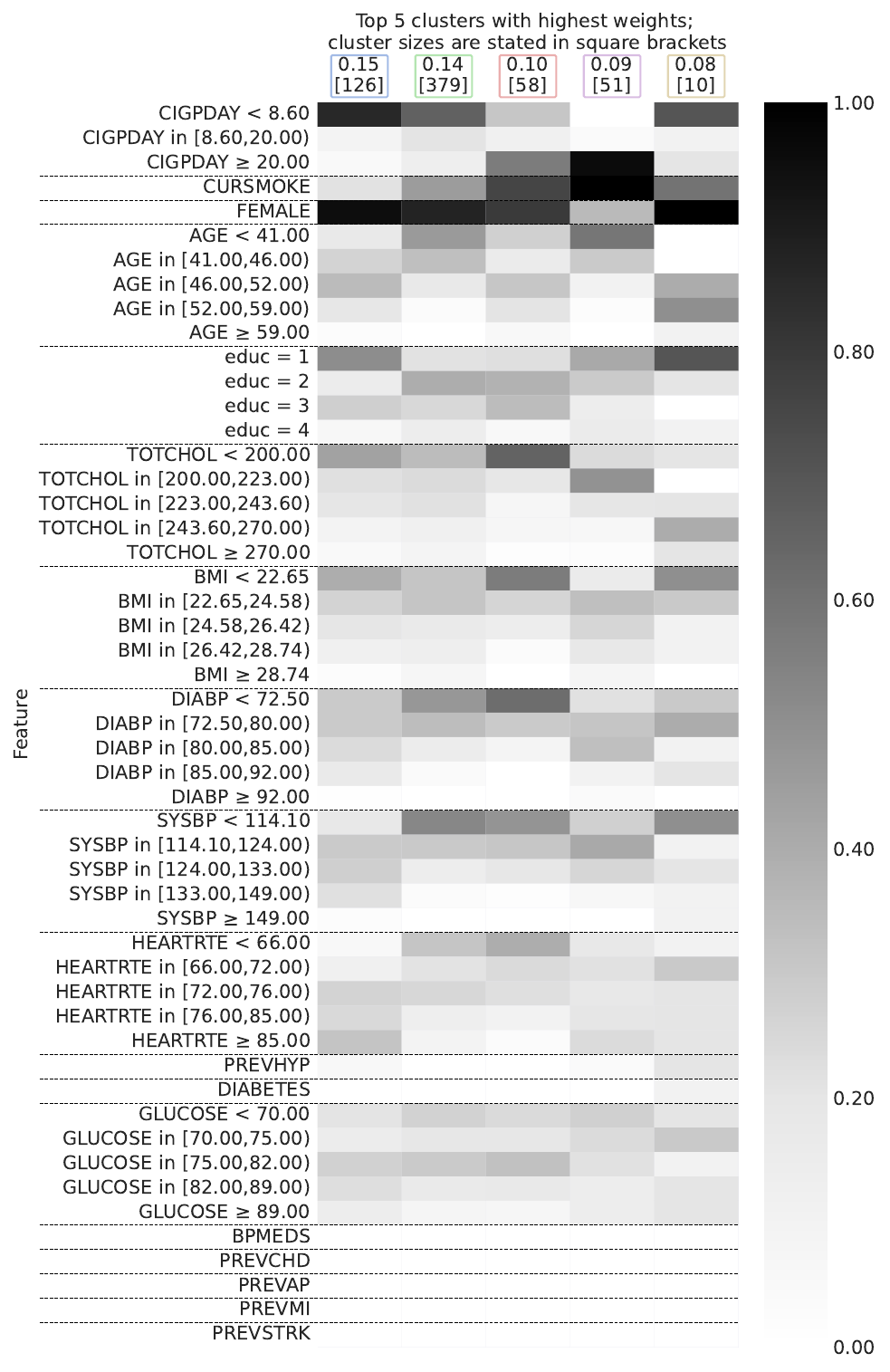}\vspace{-2em}}
\end{figure}

\paragraph{Probabilities of events happening earliest}
The probability of event $\delta\in[m]$ happening earliest for a test patient with feature vector $x$ is given by
\begin{align*}
\mathbb{P}(\Delta^*=\delta\mid X=x)
&=\mathbb{P}(\Delta^*=\delta,T\le\infty\mid X=x)\\
&=F_\delta(\infty|x),
\end{align*}
i.e., the CIF for event $\delta$ evaluated at time $\infty$.

In practice, we do not actually estimate the CIFs arbitrarily accurately all the way to time $\infty$ and can instead just extrapolate from the maximum observed time in the training data ($t_\text{max}$). For example, we can extrapolate using forward-filling (i.e., just approximate $F_\delta(\infty|x)$ with $F_\delta(t_{\max}|x)$). However, in doing so, when we sum the CIFs of the different events each evaluated at time $t_{\max}$, we are not guaranteed to get~1. We simply re-normalize to get the distribution to sum to 1 as to approximate the probability of each event occurring earliest. In other words, we approximate the probability of event $\delta$ happening earliest for feature vector $x$ by
\[
\mathbb{P}(\Delta^*=\delta\mid X=x)
\approx
\frac{F_\delta(t_{\max}|x)}{\sum_{\delta'=1}^m F_{\delta'}(t_{\max}|x)}.
\]
For example, for the same random test patient that appears in Figures~\ref{fig:individual-cif-framingham-top-5} and~\ref{fig:individual-heatmap-framingham-top-5}, at $t_{\text{max}}$, the predicted CIFs for CVD death and non-CVD death are 0.0611 and 0.0808, respectively. The probability that CVD death occurs earliest is then estimated as $\frac{0.0611}{0.0611 + 0.0808}=43.04\%$, while the probability of non-CVD death occurring first is $\frac{0.0808}{0.0611 + 0.0808}=56.96\%$.

\paragraph{Median time until an event happens given that the event is the earliest to happen}
It is sometimes also of interest to compute the median time until a given event happens for a data point. However, in the competing risks setup when $m>1$, for a particular data point, it could be that a specific event cannot happen (i.e., the data point's median time until the event is $\infty$). Instead, we can look at the median time until a given event $\delta$ happens, \emph{conditioned on event $\delta$ being the earliest to happen for a test patient with feature vector $x$}. To estimate this, note that the CDF of the time until event $\delta$ happens, conditioned on $\delta$ being the earliest event to happen for feature vector $x$ is
\begin{align*}
&\mathbb{P}(T\le t\mid X=x,\Delta^{*}=\delta)\\
&\quad=\frac{\mathbb{P}(T\le t,\Delta^{*}=\delta\mid X=x)}{\mathbb{P}(\Delta^{*}=\delta\mid X=x)}\\
&\quad=\frac{\mathbb{P}(T\le t,\Delta^{*}=\delta\mid X=x)}{\mathbb{P}(T\le\infty,\Delta^{*}=\delta\mid X=x)}\\
&\quad=\frac{F_{\delta}(t|x)}{F_{\delta}(\infty|x)}\\
&\quad\approx \frac{F_{\delta}(t|x)}{F_{\delta}(t_{\max}|x)}.
\end{align*}
The time $t$ at which this CDF crosses 1/2 is a median time estimate. Returning to the same test patient as in Figures~\ref{fig:individual-cif-framingham-top-5} and~\ref{fig:individual-heatmap-framingham-top-5}, by approximately finding $t$ for which $\frac{F_{\delta}(t|x)}{F_{\delta}(t_{\max}|x)}$ crosses 1/2 for each event $\delta$, we estimate the median time until CVD death assuming the CVD death is the earliest to happen as 2,888 days for the test subject. For non-CVD death, we instead get 2,562 days.

\subsection{Additional Cluster-Level Visualizations} \label{apd:visualization-additional-cluster-level}
We provide supplemental cluster-level views to contextualize the groups discovered by DKAJ. Specifically, we (i) present cluster-wise CIFs and feature heatmaps for PBC and SEER, and (ii)  visualize the similarity structure via a kernel-similarity heatmap for the Framingham dataset. Unless noted, clusters are ranked by size.

\paragraph{Cluster visualization for PBC and SEER}
In Section~\ref{sec:experiments}, we illustrated cluster-level visualizations using the Framingham dataset. We now provide similar analyses for the PBC and SEER datasets.

\begin{figure}[htbp]
\floatconts
  {fig:cluster-cif-pbc-top-5}
  {\caption{(PBC) CIFs for the largest 5 clusters (we then sort these 5 clusters in decreasing order by the estimated probability of death happening within the maximum observed time). Clusters correspond across the two plots in this figure as well as in Figure~\ref{fig:cluster-heatmap-pbc-top-5}.}}
  {\includegraphics[width=1.\linewidth]{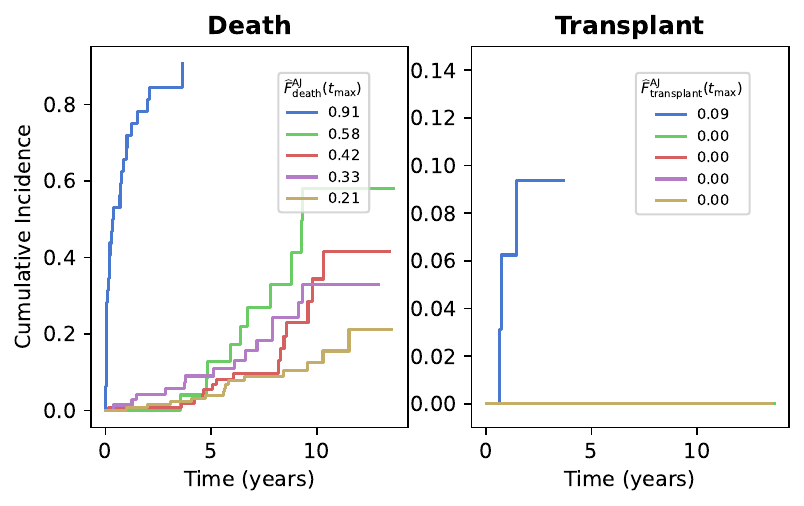}\vspace{-2.5em}}
\end{figure}

\figureref{fig:cluster-cif-pbc-top-5,fig:cluster-heatmap-pbc-top-5} present the largest five clusters identified in the PBC dataset. The CIFs in Figure~\ref{fig:cluster-cif-pbc-top-5} reveal distinct risk trajectories. For instance, the blue cluster (estimated probability of death = 0.91) exhibits both a high risk of death and an elevated probability of receiving a liver transplant. In contrast, the green cluster (risk of death = 0.58) is also associated with elevated mortality but primarily at later times, and with substantially lower likelihood of transplant compared to the blue cluster. The heatmap in Figure~\ref{fig:cluster-heatmap-pbc-top-5} highlights the covariate patterns underlying these clusters, showing that clinical characteristics such as edema indicator and histologic stage differentiate high-risk from lower-risk groups.

\begin{figure}[t!]
\floatconts
  {fig:cluster-heatmap-pbc-top-5}
  {\caption{(PBC) Feature heatmap summarizing distributions of variables in the same 5 clusters as in Figure~\ref{fig:cluster-cif-pbc-top-5}. Darker shades mean higher feature values or frequencies.}}
  {\includegraphics[width=1.\linewidth]{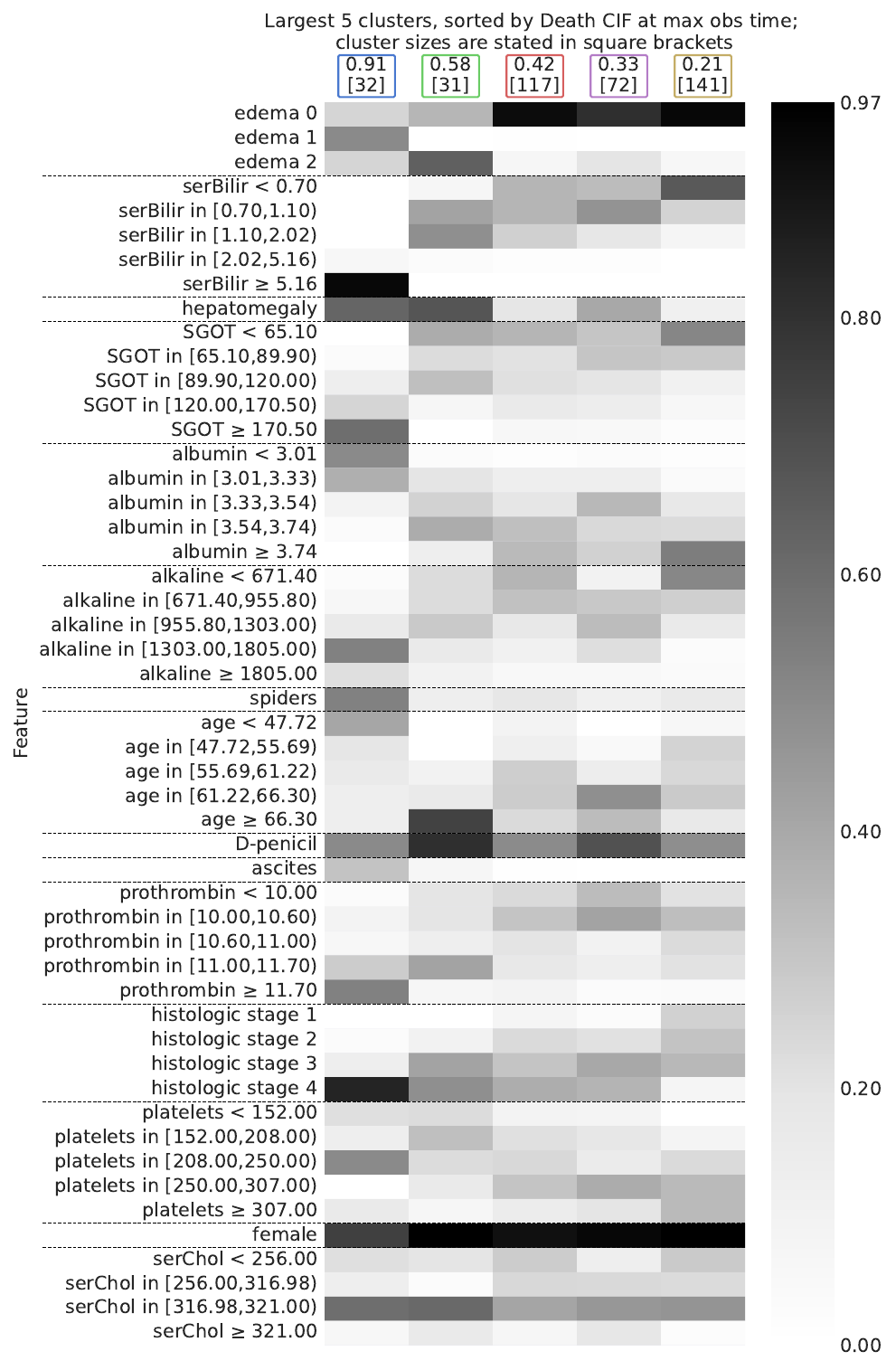}\vspace{-2.5em}}
\end{figure}

\begin{figure}[t!]
\floatconts
  {fig:cluster-cif-seer-top-5}
  {\caption{(SEER) CIFs for the largest 5 clusters (we then sort these 5 clusters in decreasing order by the estimated probability of BC happening within the maximum observed time). Clusters correspond across the two plots in this figure as well as in Figure~\ref{fig:cluster-heatmap-seer-top-5}.}}
  {\includegraphics[width=1.0\linewidth]{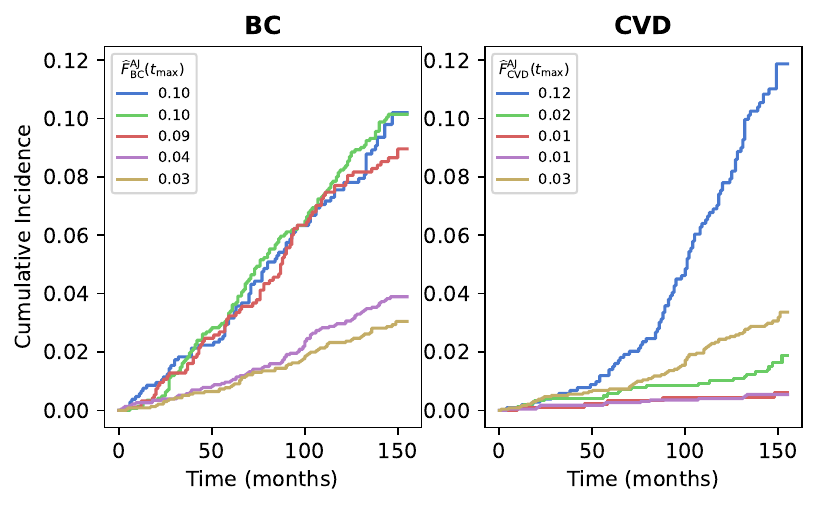}\vspace{-2.5em}}
\end{figure}

Turning to the SEER dataset, \figureref{fig:cluster-cif-seer-top-5,fig:cluster-heatmap-seer-top-5} display the largest 5 clusters of female breast cancer patients diagnosed in 2010. Here, clusters with similar breast cancer mortality (e.g., red and green versus blue) diverge markedly in their competing risk of cardiovascular death. In particular, the blue cluster shows a substantially higher probability of CVD mortality despite having death from breast cancer risk comparable to other clusters. Inspection of the feature heatmap in Figure~\ref{fig:cluster-heatmap-seer-top-5} reveals that this cluster corresponds to an older patient subgroup, consistent with the increased CVD burden.

\begin{figure*}[t!]
\floatconts
  {fig:cluster-heatmap-seer-top-5}
  {\caption{(SEER) Feature heatmap summarizing distributions of variables in the same 5 clusters as in Figure~\ref{fig:cluster-cif-seer-top-5}. Darker shades mean higher feature values or frequencies. We only display a subset of features due to space limit}}
  {\includegraphics[width=.7\linewidth]{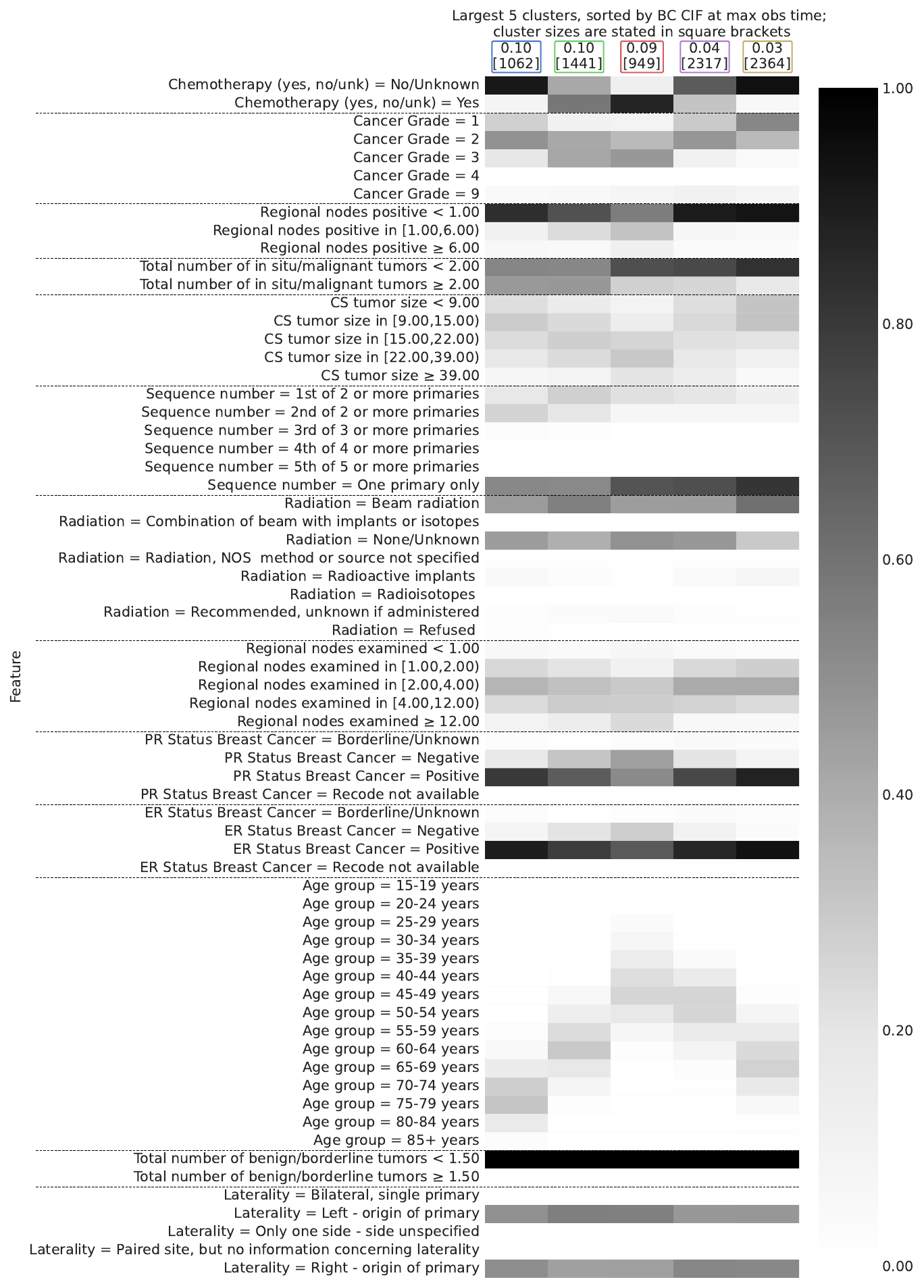}\vspace{-2.5em}}
\end{figure*}

\paragraph{Kernel similarity matrix for Framingham} 
Figure~\ref{fig:kernel-matrix-top50-clust-framingham} visualizes the learned kernel
from equation~\eqref{eq:kernel-function} after training. Rows/columns are permuted so that training points are grouped by the exemplar-based clusters produced in Step~3 of Section~\ref{sec:DKAJ-overall} (\(\varepsilon\)-net clustering). We display the 50 largest clusters for the Framingham dataset. The heatmap exhibits a clear block-diagonal pattern: high-intensity diagonal blocks indicate strong within-cluster similarity, while low off-diagonal values reflect separation across clusters; faint bands between blocks suggest clusters that are nearby in the learned embedding.

We point out that this kernel matrix can be used to help understand similarity structure using, for instance, hierarchical clustering (which could show which specific clusters of data points are considered more similar to each other according to the learned similarity/kernel matrix). In particular, many standard hierarchical clustering methods can take as input a similarity/kernel matrix instead of feature vectors (as concrete examples, the agglomerative clustering and spectral clustering methods implemented in scikit-learn \citep{pedregosa2011scikit} support this).

\begin{figure*}[t!]
\floatconts
  {fig:kernel-matrix-top50-clust-framingham}
  {\caption{Kernel similarity matrix \(K(X_i,X_j)\) (equation~\eqref{eq:kernel-function}) for the 50 largest DKAJ clusters on the Framingham dataset. Rows/columns are ordered by exemplar-based cluster assignments; darker indicates higher similarity.}\vspace{-1.1em}}
  {\vspace{-.5em}\includegraphics[width=.8\linewidth]{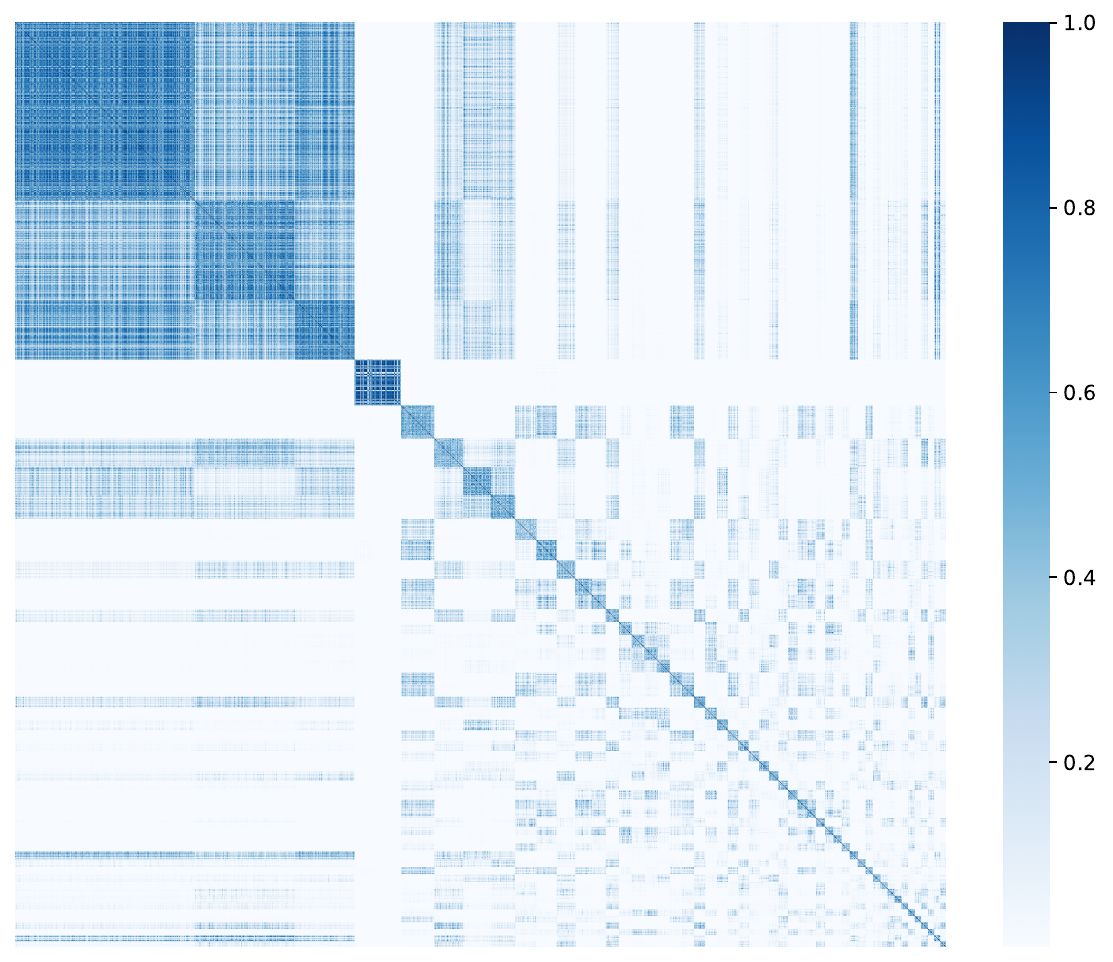}\vspace{-2.5em}}
\end{figure*}

\end{document}